\documentclass[journal,onecolumn]{IEEEtran}
\usepackage{xcolor}
\usepackage{cite}
\usepackage{amsmath,amssymb,amsfonts,amsthm}
\usepackage{graphicx}
\usepackage{multirow}
\usepackage{subfigure}
\usepackage{textcomp}
\usepackage{enumerate,paralist}
\usepackage{mathrsfs} 
\usepackage{setspace}
\usepackage{nomencl}
\usepackage{algpseudocode,algorithm}
\usepackage{threeparttable}
\usepackage{comment}
\makenomenclature

\theoremstyle{definition}

\newtheorem{lemma}{\bf Lemma}
\newtheorem{remark}{\bf Remark}

\newtheorem{corollary}{Corollary}

\graphicspath{{Graphics/}}	
\usepackage[colorlinks, linkcolor=black, anchorcolor=black, citecolor=black]{hyperref}
\usepackage{soul}

\begin{document}
\title{Improved Extended Kalman Filter-Based Disturbance Observers for Exoskeletons}
\author{
	\vskip 1em
	Shilei Li, Dawei Shi, Makoto Iwasaki, Yan Ning, Hongpeng Zhou, Ling Shi
	
	\thanks{Shilei Li and Dawei Shi are with School of Automation, Beijing Institute of Technology, Beijing 100081, China (e-mail: shileili@bit.edu.cn, daweishi@bit.edu.cn)}
	\thanks{Makoto Iwasaki is with the Department of Electrical and Mechanical
		Engineering, Nagoya Institute of Technology, Nagoya 466-8555, Japan
		(e-mail: iwasaki@nitech.ac.jp)}
	\thanks{Yan Ning and Ling Shi are with Department of Electronic and Computer Engineering, Hong Kong University of Science and Technology, Hong Kong SAR (e-mail: yningaa@connect.ust.hk, eesling@ust.hk)}
	\thanks{Hongpeng Zhou is with Department of Computer Science, University of Manchester, Manchester, United Kingdom (e-mail: hongpeng.zhou@manchester.ac.uk)}
}
	\maketitle

\begin{abstract}
	The nominal performance of mechanical systems is often degraded by unknown disturbances. A two-degree-of-freedom control structure can decouple nominal performance from disturbance rejection. However, perfect disturbance rejection is unattainable when the disturbance dynamic is unknown. In this work, we reveal an inherent trade-off in disturbance estimation subject to tracking speed and tracking uncertainty. Then, we propose two novel methods to enhance disturbance estimation: an interacting multiple model extended Kalman filter-based disturbance observer and a multi-kernel correntropy extended Kalman filter-based disturbance observer. Experiments on an exoskeleton verify that the proposed two methods improve the tracking accuracy $36.3\%$ and $16.2\%$ in hip joint error, and $46.3\%$ and $24.4\%$ in knee joint error, respectively, compared to the extended Kalman filter-based disturbance observer,  in a time-varying interaction force scenario, demonstrating the superiority of the proposed method. 
\end{abstract}

\begin{IEEEkeywords}
	Disturbance observer, multi-kernel correntropy, interacting multiple model, bias-variance tradeoff, human-robot interaction
\end{IEEEkeywords}

{}

\definecolor{limegreen}{rgb}{0.2, 0.8, 0.2}
\definecolor{forestgreen}{rgb}{0.13, 0.55, 0.13}
\definecolor{greenhtml}{rgb}{0.0, 0.5, 0.0}

\section{Introduction}

\IEEEPARstart{D}{isturbances} are common in various robotic systems, including robotic arms~\cite{bb1}, unmanned aerial vehicles~\cite{bb2,cao2025proximal}, and aircraft~\cite{bb3}, where they degrade system performance and can even cause instability~\cite{bb4}. In exoskeletal robots, disturbances arise from unmodeled joint friction, unknown loads, and human–robot interactions. If not properly addressed, these unforeseen forces can significantly impair the exoskeleton’s control performance.

\textcolor{black}{A common approach to mitigating disturbances is through the design of robust controllers. However, in traditional single-degree-of-freedom control structures, there exists an inherent balance between nominal performance and robustness~\cite{bb4}. To overcome this limitation, a promising technique involves incorporating disturbance observers (DOBs). By introducing DOBs into the system, a two-degree-of-freedom control configuration is achieved, where DOBs complement the controller and allow for enhanced system robustness without compromising nominal performance.} Various disturbance observers have been developed for different applications, including the nonlinear disturbance observer (NDOB)\cite{bb5}, extended state observer (ESO)\cite{bb6}, equivalent input disturbance (EID)~\cite{bb30}, generalized proportional–integral observer (GPIO)\cite{bb7}, unknown input observer (UIO)\cite{bb8}, simultaneous input and state estimator (SISE)\cite{bb9}, and Kalman filter-based disturbance observer (KF-DOB)\cite{bb10}. \textcolor{black}{Recently, some data-driven approaches have emerged, which explore the mapping between the measurements and the disturbance. Bisheban et al.~\cite{9143160} utilized a neural network to approximate the unknown disturbances by leveraging the universal approximation theorem. Lutter et al.~\cite{lutter2019deep} developed a Deep Lagrangian Networks (DeLaN) for unknown dynamics learning by exploiting the Euler-Lagrange equation. To further enhance the system's performance against disturbance, meta-learning-based methods have been developed to achieve agile and precise quadrotor flight tracking with different wind conditions~\cite{o2022neural,wei2025meta}. Although these methods achieve satisfactory performance, they are limited to specific disturbances and need to be trained carefully. In exoskeletons, the disturbances partially come from the active intention of subjects, which is personalized and is less structural, posing challenges to the learning-based method. In this aspect, the conventional disturbance observers are advantageous.} Notably, observers, such as NDOB, ESO, GPIO, and UIO, overlook the influence of stochastic noise within the system. In contrast, the Kalman filter offers a more comprehensive framework, providing both state estimation and associated uncertainty quantification. Hence, this work focuses on the KF-based disturbance observers.

In our previous work~\cite{bb29}, we demonstrated that disturbance estimation using the Kalman filter (KF) with stochastic noise and \emph{incomplete disturbance models} involves an inherent trade-off. In this study, we extend this finding to the nonlinear domain, specifically for the extended Kalman filter (EKF), showing that two key objectives—tracking speed (i.e., bias) and tracking smoothness (i.e., variance)—are fundamentally conflicting under fixed observer parameters. Although not explicitly proved except in work~\cite{bb29}, this phenomenon has been observed in many practical implementations. For example, in high-gain observers, increased gains yield faster convergence but cause significant overshoot during the initial phase, commonly known as the peaking phenomenon~\cite{bb26}. Similarly, in NDOBs~\cite{bb5}, a larger bandwidth improves convergence speed but also amplifies noise effects. A typical strategy to manage this trade-off is careful parameter tuning. However, this process is time-consuming and often yields unsatisfactory results. In exoskeletons, the coexistence of both fast-varying disturbances (e.g., friction or human intention) and slow-varying disturbances makes parameter tuning challenging.

Switched or adaptive filter gain has the potential to address the aforementioned challenges. In this study, we first show that the performance of a two-degree-of-freedom control structure is affected by the accuracy of disturbance estimation, as demonstrated using a Lyapunov function analysis. We then reveal that the extended Kalman filter-based disturbance observer (EKF-DOB) is unsatisfactory for scenarios with both fast and slow varying disturbances. To address this issue, we propose two solutions: the interacting multiple model extended Kalman filter-based disturbance observer (IMMEKF-DOB) and the multi-kernel correntropy extended Kalman filter-based Disturbance Observer (MKCEKF-DOB). The former mitigates the bias-variance dilemma by employing a switched process noise covariance, while the latter achieves similar performance through multi-kernel correntropy and fixed-point iteration. Different from the conventional interacting multiple model Kalman filter (IMMKF), which builds upon  Markov jump systems and switches between different models~\cite{bb13,bb14,bb15,bb16,bb17}, we introduce a multiple disturbance covariance scheme, which dynamically adjusts observer parameters based on the innovation characteristics. The multi-kernel correntropy Kalman filter (MKCKF) was introduced in our previous work~\cite{bb19}. Here, we extend this to a nonlinear version that provides an ``adaptive covariance" through the multi-kernel correntropy properties~\cite{bb19}. \textcolor{black}{It is worth mentioning that the main purpose of this work is to provide alternatives to EKF-DOB with complex disturbances, rather than providing the state-of-the-art exoskeleton angle tracking performance.} The contributions of this work are summarized as follows:
\begin{itemize}
	\item We demonstrate that there exists an inherent bias-variance trade-off in EKF-DOB with fixed process covariance, under the assumption that the disturbance dynamics are partially unknown, which impedes its application under complex disturbance scenarios.
	\item We developed two novel algorithms, i.e., IMMEKF-DOB and MKCEKF-DOB, which provide a better bias-variance than the conventional EKF-DOB, and hence provide better disturbance estimation and tracking performance. 
	\item We validate the effectiveness of the proposed algorithms through both extensive simulations and experiments with open-sourced code at \url{https://github.com/lsl-zsj/ImprovedEKF-DOB}.
\end{itemize}

The remainder of this paper is arranged as follows. In Section II, we provide the exoskeleton models and give the problem statement. In Section III, we develop two novel observers, i.e., IMMEKF-DOB and MKCEKF-DOB. In Section IV, we conduct some simulations and experiments. In Section V, we draw a conclusion.

\emph{Notations}: The transpose of a matrix $A$ is denoted by $A^{T}$. The notation $X\succ 0$ ($X \succcurlyeq 0$) indicates that $X$ is a positive definite (semi-positive definite) matrix. The symbol $\|x\|_{A}$ denotes $x^{T}Ax$. The notation $A \to \infty$ denotes that all eigenvalues of $A$ tend to infinity, which implies that $A^{-1} \to 0$.

\section{Problem Statement}
In this section, we first present the system model of the exoskeleton. Next, we demonstrate that the performance of the disturbance observer determines the system’s error bound for a given controller. Finally, we give the problem statement.
\subsection{System Modeling}
\begin{figure}[htbp]
	\centering
	\subfigure[]{\includegraphics[width=1.4in]{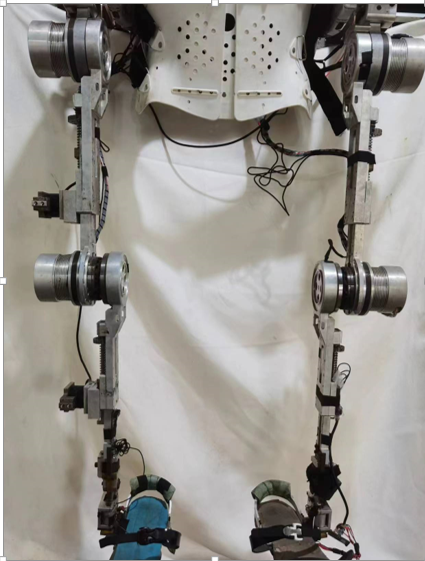}%
		\label{exo}}
	\hfil
	\subfigure[]{\includegraphics[width=1.4in]{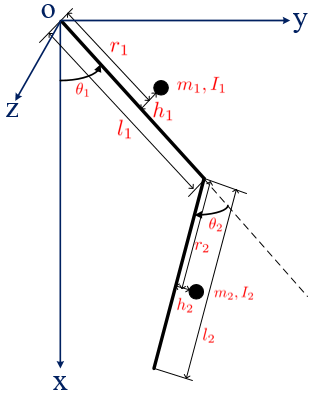}%
		\label{exo_link}}
	\caption{The exoskeleton and the diagram of two-link robotic leg model. The symbol $m$ is the mass, $I$ is the inertia, $l$ is the link length, $r$ is the distance from joints to the center of mass along the direction of the link, $h$ is the distance from the center of mass to the link, and $\theta$ is the rotation angle. The subscript $1,2$ represents the corresponding values for link 1 and link 2.}		
	\label{EXO}
\end{figure}

The exoskeleton is shown in Fig. \ref{exo}. In our applications, the exoskeleton's waist is fixed to a frame. Hence, each leg of the exoskeleton is modeled as a two-link robotic leg. The exoskeleton's schematic diagram is shown in Fig. \ref{exo_link}. 

According to the Euler-Lagrange equation~\cite{bb20}, the exoskeleton dynamics follows
\begin{equation}
	M(\theta)\ddot{\theta} + C(\theta,\dot{\theta})\dot{\theta} + N(\theta) = \tau + d 
	\label{sysdyn}
\end{equation}
where $M(\theta) \in \mathbb{R}^{2}$ is the positive definite inertial matrix, $C(\theta,\dot{\theta}) \in \mathbb{R}^{2}$ is the Coriolis matrix, and $N(\theta) \in \mathbb{R}^{2}$ is the gravity matrix. The system identification procedure and system parameters are available in \textcolor{black}{Section \ref{abc}}. 
\subsection{Stability Analysis of the Augmented PD Controller}
The augmented proportional-derivative (PD) controller is the most popular controller in industries~\cite{bb20}, but its performance is affected by the unknown disturbances. To reject the disturbance, a disturbance observer is designed, which forms the following two-degree-of-freedom control system configuration:
\begin{equation}
	\begin{aligned}
		\tau&=\tau_{ff}+\tau_{fb} - \hat{d}\\
		\tau_{ff}&=M(\hat{\theta}) \ddot{\theta}_d+C(\hat{\theta}, \hat{\dot{\theta}}) \dot{\theta}_d+N(\hat\theta)\\
		\tau_{fb}&=-K_d (\hat{\dot\theta}-\dot{\theta}_d)-K_p (
		\hat{\theta}-\theta_d)
		\label{controller}
	\end{aligned}
\end{equation}
where $\tau_{ff}$ is the feedforward torque, $\tau_{fb}$ is the feedback torque, and $\hat{d}$, $\hat{\theta}$, and $\hat{\dot\theta}$ are corresponding estimates of disturbance and states. Denote $\tilde{d}_e=\hat{d}-d$, $\tilde{\theta}_e=\theta-\hat{\theta}$, and $\tilde{\dot\theta}_e=\dot\theta-\hat{\dot\theta}$. Let  $e=\theta-\theta_d$ and $\dot{e}=\dot\theta-\dot{\theta}_d$. Substituting  \eqref{controller} into \eqref{sysdyn} arrives
\begin{equation}
	M(\theta) \ddot{e}+C(\theta, \dot{\theta}) \dot{e}+K_d (\dot{e}-\tilde{\dot\theta}_e)+K_p (e-\tilde{\theta}_e) -\tilde{d}_e-\delta=0.
	\label{closesys1}
\end{equation}
where $\delta= \tau_{ff}-\tau_{ff}^{*}=\big(M(\hat{\theta})-M({\theta})\big) \ddot{\theta}_d+\big(C(\hat{\theta}, \hat{\dot{\theta}})-C({\theta}, {\dot{\theta}}) \big)\dot{\theta}_d+N(\hat\theta)-N(\theta)$ is the feedforward error. Denoting the lumped error as $l_{e}=-K_d \tilde{\dot\theta}_e - K_p \tilde{\theta}_e -\tilde{d}_e - \delta$ gives
\begin{equation}
	M(\theta) \ddot{e}+C(\theta, \dot{\theta}) \dot{e}+K_d \dot{e}+K_p e +l_e=0.
	\label{closesys}
\end{equation}

\begin{lemma}[\hspace{1sp}\cite{bb20}]
	If $l_e \triangleq 0$, the closed system \eqref{closesys} is exponentially stable as long as $K_p \succ 0$ and $K_d \succ 0$. 
\end{lemma}

\textcolor{black}{In practical applications, perfect disturbance and state estimation are unattainable and there is a bias-variance tradeoff under unknown disturbance dynamics in the KF-DOB (see Theorem 3 of ~\cite{bb29}). The conclusion can be shifted to the EKF-DOB, which reveals that the disturbance tracking speed (i.e., the estimation bias) and the disturbance smoothness (i.e., the estimation variance) are two conflicting goals with respect to tunable covariance parameters.} The following lemma describes how the lumped disturbance error influences the stability of the system.
\begin{lemma}
	The system \eqref{closesys} is globally uniformly ultimately bounded for $K_d \succ 0$ and $K_p \succ 0$ with bounded $l_e$. 
	\label{lemma2}
\end{lemma}
The proof of this lemma is available in Appendix \ref{appen1}. 
\textcolor{black}{
	\begin{corollary}
		\label{corollary1}
		Given the bounded lumped disturbance error $\|l_e\| \leq \bar{l}_e$, the closed-loop system converges to the ellipsoidal invariant set:
		\begin{equation}
			\mathcal{B} = \left\{ (e, \dot{e}) : \alpha_1 \|e\|^{2} + \alpha_2 \|\dot{e}\|^{2} \leq \kappa \right\}
			\label{region}
		\end{equation}
		where the convergence radius $\kappa$ is defined as
		\[
		\kappa = \dfrac{\bar{l}_e^2}{2} \left( \dfrac{1}{\lambda_{\min}(K_d)} + \dfrac{\epsilon}{\lambda_{\min}(K_p)} \right),
		\] $\alpha_1$, $\alpha_2$, and $\epsilon$ are constants shown in equation \eqref{expconv}, and $\lambda_{\min}(\cdot)$ denotes the minimum eigenvalue of a positive definite matrix.
\end{corollary}}
The proof of the corollary is available in Appendix \ref{appen2}.
\textcolor{black}{
	\begin{remark}
		An intuitive strategy of shrinking the invariant set is to increase the PD gains. However, this approach may cause the lumped disturbance $l_e$ to become unbounded due to the amplification of angle and angular velocity estimation errors $\tilde{\theta}_e$ and $\tilde{\dot\theta}_e$. In such situations, one may resort to much more accurate lumped disturbance estimate.
\end{remark}}
\begin{remark}
	In most existing disturbance observers, a trade-off exists between disturbance tracking speed and tracking smoothness~\cite{bb5,bb29}. Specifically, in KF-DOB, the fastest tracking speed is achieved by setting an infinite disturbance covariance, at the cost of the largest estimation uncertainty( in such cases, one obtains a unbiased minimum variance estimator)~\cite{bb29,Gillijns}. Conversely, using a smaller disturbance covariance yields a smoother disturbance estimate but reduces tracking speed. Therefore, it may be beneficial to \emph{apply a switched or adaptive disturbance covariance} in a complex disturbance scenario.  
\end{remark}
\subsection{Problem Statement}
Given controller \eqref{controller}, the error invariant set is determined by the lumped disturbance according to Corollary \ref{corollary1}. A conventional approach involves designing disturbance observers, such as the EKF-DOB, to simultaneously estimate both the states and disturbances. However, we contend that EKF-DOB is inadequate for exoskeleton applications characterized by both fast and slow-varying disturbances. To address this challenge, we develop novel observers that not only mitigate this issue but also yield improved tracking performance.
\section{Methodology}
We first develop an extended Kalman filter-based disturbance observer for exoskeletons. Then, we provide two remedies, i.e., IMMEKF-DOB and MKCEKF-DOB, which provide a better trade-off between disturbance tracking speed and tracking variance. 
\subsection{Extended Kalman Filter-Based Disturbance Observer}
For exoskeletons shown in Fig. \ref{EXO} and considering the addictive noise, we construct the following state space dynamic model:
\begin{equation}
	\begin{aligned}
		\dot{x} = \mathrm{f}(x)+w\\
		y=\mathrm{g}(x)+v
	\end{aligned}
	\label{ssm}
\end{equation}
with
\begin{equation}
	\begin{aligned}
		\mathrm{f}(x)&=\begin{bmatrix}
			0\\
			\dot{\theta}\\
			M(\theta)^{-1}\big(\tau+d-C(\theta,\dot{\theta})\dot{\theta}-G(\theta)\big)
		\end{bmatrix}\\
		\mathrm{g}(x)&=\mathrm{H}x=\begin{bmatrix}
			0,0,1,0,0,0\\
			0,0,0,1,0,0\\
			0,0,0,0,1,0\\
			0,0,0,0,0,1\\
		\end{bmatrix}x,
	\end{aligned}
\end{equation}
where the augmented state $x=[{d}^{T},{\theta}^{T},\dot{\theta}^{T}]^{T} \in \mathbb{R}^{6}$ contains both the disturbance and the state, $y$ contains the angle and angular velocity measurement, $\mathrm{f}(x)$ and $\mathrm{g}(x)$ are state transfer and observer functions, and $w$ and $v$ are process and measurement noises. The nominal disturbance dynamics has $\dot{d}=0$ which follows works \cite{bb5,bb22}. Subsequently, we discretise \eqref{ssm} and obtain
\begin{equation}
	\begin{aligned}
		x_{k}&=f(x_{k-1})+w_{k-1}\\
		y_{k}&=g(x_{k})+v_{k}
		\label{discreteDyn}
	\end{aligned}
\end{equation}
where $x_k=[d_k^{T}, s_k^{T}] \in \mathbb{R}^{n}$ incorporating the disturbance $d_k \in \mathbb{R}^{p}$ and state $s_k=[\theta_k^{T},\dot{\theta}_k^{T}]^{T} \in \mathbb{R}^{n-p}$, ${f}(x_{k-1})=x_{k-1}+\mathrm{f}(x_{k-1})\Delta T$, $g(x_{k})=H_k x_k \in \mathbb{R}^{m}$ with $H_k \triangleq \mathrm{H}$, $k$ is the time index, and $\Delta T$ is the sampling interval. We denote $F_{k}=\frac{\partial f}{\partial x_k}|_{\hat{x}_{k|k}}$,$Q_k= E[w_k^{T}w_k]$, and $R_k=E[v_k^T v_k]$. Moreover, we assume disturbance noise is independent with the state noise, i.e., 
$Q_k=\begin{bmatrix}
	Q_{d,k}&0\\
	0&Q_{s,k}
\end{bmatrix}$. Then, one can execute EKF-DOB as follows:
\begin{equation}
	\begin{aligned}
		&\hat{x}_{k|k-1}={f}(\hat{x}_{k-1|k-1})\\
		&P_{k|k-1}=F_{k-1} P_{k-1|k-1} F_{k-1}^{T} + Q_{k-1}\\
		&K_k=P_{k|k-1}H_k^{T}(H_k P_{k|k-1} H_k^{T}+ R_k)^{-1}\\
		&\hat{x}_{k|k}=\hat{x}_{k|k-1}+K_{k}(y_k-H_k x_k)\\
		&P_{k|k}=(I-K_k H_k)P_{k|k-1}
	\end{aligned}
	\label{ekf-dob}
\end{equation}
where $P_{k|k}$ is the error covariance and $K_k$ is the Kalman gain.
\begin{remark}
	\textcolor{black}{The nominal disturbance model $d_k = d_{k-1} + w_{d,k-1}$ is adopted in \eqref{discreteDyn} due to the lack of prior knowledge about the disturbance dynamics.} We assume that the true disturbance evolves according to $d_k = f(d_{k-1}) + w_{d,k-1}^{*}$, where $f(d_{k-1})$ is a time-varying function and $w_{d,k-1}^{} \sim \mathcal{N}(0, Q_{d,k-1}^{})$. Substituting this into the nominal model yields
	\begin{equation}
		w_{d,k-1} = f(d_{k-1}) - d_{k-1} + w_{d,k-1}^{*},
		\label{heavydist}
	\end{equation}
	which includes both the model mismatch $f(d_{k-1}) - d_{k-1}$ and the Gaussian noise $w_{d,k-1}^{*}$. \textcolor{black}{Following the insights from Theorem 3 of ~\cite{bb29} and extending them to the nonlinear setting, assigning $Q_{d,k} \to \infty$ yields unbiased state and disturbance estimates, though at the cost of increased estimation variance (see Fig.~\ref{pdobs0} with $\eta = \exp(40)$). In contrast, setting $Q_{d,k} = Q_{d,k}^{*}$ minimizes the estimation variance when $f(d_{k-1}) \triangleq d_{k-1}$, i.e., when the nominal model matches the true disturbance dynamics, but may lead to slow convergence under rapidly changing disturbances (see Fig.~\ref{pdobs1} with $\eta = \exp(0)$).} This is called the bias-variance tradeoff, which can be mitigated by employing an adaptive process covariance, as discussed in the following section.
\end{remark}
\subsection{Interacting Multiple Model Extended Kalman Filter-Based Disturbance Observer}
As demonstrated in \eqref{heavydist}, when the disturbance varies slowly, the model mismatch $f(d_k)-d_k$ is minor and one can apply $Q_{d,k}=Q_{d,k}^{*}$. On the contrary, when disturbance varies fast,  $f(d_k)-d_k$ becomes large and one can use $Q_{d,k}=\eta Q_{d,k}^{*}$ where $\eta>1$ is a scalar which reflects the level of model mismatch. To parallel execute the same disturbance model but with different disturbance noise covariance, we  construct the IMMEKF-DOB with $j$ different disturbance process covariance as follows ($j=1,2,\ldots,p$):
\begin{equation}
	\begin{aligned}
		{x}_{k}&=f({x}_{k-1})+{w}_{j,k-1}\\
		{y}_k&=g({x}_k)+{v}_{k}
		\label{linear}
	\end{aligned}
\end{equation}
where ${w}_{j,k} \sim \mathcal{N}(0,Q_{j,k})$ is are Gaussian noises for $j$-th model with $Q_{j,k}=\begin{bmatrix}
	Q_{j,d,k}&0\\
	0&Q_{s,k}
\end{bmatrix}$. The Markov transition probability matrix has $P=[\mathcal{P}_{i,j}]$ where $\mathcal{P}_{i,j}$ is the transition probability from model $i$ to model $j$. Then, the IMMEKF-DOB is developed in Algorithm \ref{immkf-dob} by analogy with the IMMKF~\cite{bb16}.
\begin{algorithm}[htbp]
	\setstretch{0.9} 
	\caption{{IMMEKF-DOB}}
	\label{immkf-dob}
	\begin{algorithmic}[1]
		\State {\textbf{Step 1: Initialization}}\\
		Select disturbance noise covariance $\mathrm{Q}_{j,d,k}$ for model $j$.
		\State {\textbf{Step 2: Mixing}}
		\State Input interaction\\
		$\mu_{ij,k-1|k-1}=\mathcal{P}_{ij}\mu_{i,k-1}/\bar{c}_{j}$
		where $\bar{c}_{j}=\sum_{i=1}^{p}\mathcal{P}_{ij}\mu_{i,k-1}$ 
		\State Obtain initial state and covariance of model $j$\\
		$\hat{x}_{j,k-1|k-1}^{init}=\sum_{i=1}^{p}\hat{x}_{i,k-1|k-1}\mu_{ij,k-1|k-1}$\\
		$P_{j,k-1|k-1}^{init}=\sum_{i=1}^{p}\mu_{ij,k-1|k-1}\Big{(}P_{i,k-1|k-1}+\big{(}\hat{x}_{i,k-1|k-1}-\hat{x}_{i,k-1|k-1}^{init}\big{)}
		\big{(}\hat{x}_{i,k-1|k-1}-\hat{x}_{i,k-1|k-1}^{init}\big{)}^{T}\Big{)}$
		\State {\textbf{Step 3: Filtering}}\\
		$\hat{x}_{j,k|k}=\hat{x}_{j,k|k-1}+K_{j,k}e_{j,k}$\\
		$\hat{x}_{j,k|k-1}= f(\hat{x}_{j,k-1|k-1}^{init})$\\
		$P_{j,k|k-1}=F_{j,k-1}P_{j,k-1|k-1}^{init}F_{j,k-1}^{T}+ Q_{j,k-1}$\\
		$F_{j,k-1}=\frac{\partial f}{\partial x_k}|_{\hat{x}_{j,k-1|k-1}^{init}}, H_{j,k}=\frac{\partial g}{\partial x_k}|_{\hat{x}_{j,k|k-1}}$\\
		$e_{j,k}= y_k-g(\hat{x}_{j,k|k-1})$\\
		$S_{j,k}=H_{j,k} P_{j,k|k-1} H_{j,k}^{T}+R_{k}$\\
		$K_{j,k}=P_{j,k|k-1}H_{j,k}^{T}S_{j,k}^{-1}$\\
		$P_{j,k|k}=\big{(}I-K_{j,k}H_{j,K}\big{)}P_{j,k|k-1}$
		\State {\textbf{Step 4: Model Probability Update}}\\
		$\mu_{j,k}=\frac{\Lambda_{j,k}\bar{c}_{j}}{c}$ \\
		$			\Lambda_{j,k}=\frac{1}{\sqrt{2\pi|S_{j,k}|}}\exp\big{(}-\frac{1}{2}e_{j,k}^{T}S_{j,k}^{-1}e_{j,k}\big{)}$\\
		$c=\sum_{j=1}^{p}\Lambda_{j,k}\bar{c}_{j} $
		\State {\textbf{Step 5: Output}}\\
		$\hat{x}_{k|k}=\sum_{j=1}^{p}\mu_{j,k}\hat{x}_{j,k|k}$ \\
		$P_{k|k}=\sum_{j=1}^{p}\mu_{j,k}{(}P_{j,k|k}+{(}\hat{x}_{j,k|k}-\hat{x}_{k|k}{)}{(}\hat{x}_{j,k|k}-\hat{x}_{k|k}{)}^{T}{)}.$
	\end{algorithmic}
\end{algorithm}
\subsection{Multi-Kernel Correntropy Extend Kalman Filter-based Disturbance Observer}
In our previous work~\cite{bb19}, we coined the terminology \emph{multi-kernel correntropy} (MKC) which is less conservative than the \emph{correntropy} ~\cite{bb23}. The MKC is a similarity measure of two random vectors $\mathscr{X}, \mathscr{Y} \in \mathbb{R}^{l}$: 
\begin{equation}
	V(\mathscr{X},\mathscr{Y})= \sum_{i=1}^{l} \sigma_i^2 E[ \kappa_{\sigma_i}(\mathscr{X}_i,\mathscr{Y}_i)]
\end{equation}
where $E[\kappa_{\sigma_i}(\mathscr{X}_i,\mathscr{Y}_i)]=\int \kappa_{\sigma_i}(x_i,y_i)d F_{\mathscr{X}_i\mathscr{Y}_i}(x_i,y_i)$, $\kappa_{\sigma_i}(x_i,y_i)$ is a shift-invariant Mercer Kernel. We use Gaussian kernel $\kappa_{\sigma_i}(x_i,y_i)=G_{\sigma_i}(x_i,y_i)=\exp(-\frac{e_i^2}{2\sigma_i^2})$ where $\sigma_i$ is the kernel bandwidth and $e_i=x_i-y_i$ in this work. In some applications, only finite samples $x_{k}$ and $y_{k}$ can be obtained. In such cases, the MKC loss (MKCL) is defined as 
\begin{equation}
	\begin{aligned}
		J_{MKCL}&=\sum_{i=1}^{l}\sigma_i^2 (1-\hat{V}_{i})=\frac{1}{N}\sum_{k=1}^{N}\sum_{i=1}^{l}\sigma_i^2 \Big(1-G_{\sigma_i}(e_{i,k})\Big)
		\label{GL}
	\end{aligned}
\end{equation}
where $e_k=[e_{1,k},e_{2,k},\ldots,e_{l,k}]^{T}$ and $e_{i,k}=x_{i,k}-y_{i,k}$. 

We rewrite \eqref{discreteDyn} as (note that $g(x_k)=H_k x_k$)
\begin{equation}
	{T}_k={W}_k {x}_k + {B}_{k}^{-1}{\nu}_k
\end{equation}
where 
\begin{equation}\nonumber
	{T}_k={B}_{k}^{-1}\begin{bmatrix}
		{\hat{x}}_{k|k-1}\\
		{y}_k
	\end{bmatrix},~
	{W}_k={B}_{k}^{-1}\begin{bmatrix}
		{I}\\
		{H}_k
		\label{TW}
	\end{bmatrix},~{\nu}_k=\begin{bmatrix}
		{\hat{x}}_{k|k-1}-{x}_k\\
		{v}_k
	\end{bmatrix}\\
\end{equation}
and ${B}_{k}$ is obtained by Cholesky decomposition with 
\begin{equation}
	\small
	\begin{aligned}
		E({\nu}_k {\nu}_k^{T})&=\begin{bmatrix}
			{P}_{k|k-1}&0\\
			0&{R_k}
		\end{bmatrix}=\begin{bmatrix}
			{B}_{p}{B}_{p}^{T}&0\\
			0&{B}_{r}{B}_{r}^{T}
		\end{bmatrix}={B}_{k}{B}_{k}^{T}.
	\end{aligned}
	\label{bpbr}
\end{equation}
One observes that $e_k={T}_k-{W}_k {x}_k={B}_{k}^{-1}{\nu}_k \sim \mathcal{N}(0,I)$. The conventional EKF-DOB can be obtained by solving the following LS cost function~\cite{bb19}
\begin{equation}
	\min_{\mathrm{x}_k} J_{LS}= \frac{1}{2}\|e_{k}\|_2^{2}.
	\label{ekf}
\end{equation}
As a comparison, the MKCEKF-DOB is obtained by solving
\begin{equation}
	\min_{\mathrm{x}_k} J_{MKCEKF} =  \sum_{i=1}^{l} \sigma_{i}^{2}\big(1-G_{\sigma_{i}}({e}_{i,k})\big)
	\label{mkcekf}
\end{equation}
where ${e}_{i,k}$ is the $i$-th element of $e_k$. 
\begin{lemma}[~\cite{bb25}]
	The MKCL in \eqref{mkcekf} is identical to the LS criterion in \eqref{ekf} as all $\sigma_i \to \infty$.
\end{lemma}
Denote the residual as ${e}_k=[e_{p,k}^{T},e_{r,k}^{T}]^{T}=[{e}_{d,k}^{T},{e}_{s,k}^{T},e_{r,k}]^{T}$ where $e_{p,k}$ is the process error, $e_{r,k}$ is the measurement error, ${e}_{d,k}$ is the disturbance error, and ${e}_{s,k}^{T}$ is the state error. Moreover, we denote the kernel bandwidth vector as $\boldsymbol{\sigma}=[\sigma_{p}^{T},\sigma_r^{T}]^{T}=[\sigma_{d}^{T},\sigma_{s}^{T},\sigma_{r}^{T}]^{T}$. According to \eqref{heavydist}, only the disturbance channel is heavy-tailed due to the existence of model mismatch. Based on the mapping between the objective function and the induced probability density function shown in \cite{bb19}, we apply $\sigma_{s} \to \infty$ and $\sigma_{r}\to \infty$, which gives
\begin{equation}
	\begin{aligned}
		\min_{\mathrm{x}_k} J_{MKCDOB} &=  \sum_{i=1}^{p} \sigma_{d,i}^{2}\big(1-G_{\sigma_{d,i}}({e}_{d,i,k})\big) \\
		&+ \frac{1}{2}\|e_{s,k}\|_2^{2} +\frac{1}{2}\|e_{r,k}\|_2^{2} 
		\label{mkcekfdobl}
	\end{aligned}
\end{equation}
where $\sigma_{d,i}$ is the $i$-th element of $\sigma_{d}$ and ${e}_{d,i,k}$ is $i$-th element of ${e}_{d,k}$. After a similar derivation shown in \cite{bb19} and extend the corresponding result to the nonlinear domain by applying first-order Taylor expansion, we obtain Algorithm \ref{mkcekfdob}. 
\begin{algorithm}
	\setstretch{1.0} 
	\caption{{MKCEKF-DOB}}
	\label{mkcekfdob}
	\begin{algorithmic}[1]
		\State {\textbf{Step 1: Initialization}}\\
		Choose ${\sigma}_d$ and a threshold $\varepsilon$
		\State {\textbf{Step 2: State Prediction}}\\
		$\hat{{x}}_{k|k-1}= f({x}_{k-1|k-1}) $, $F_{k-1}=\frac{\partial f}{\partial x_k}|_{\hat{x}_{k-1|k-1}}$\\
		${P}_{k|k-1}={F_{k-1}} {P}_{k-1|k-1 } {F_{k-1}}^{T}+{Q}_{k-1}$
		\State {\textbf{Step 3: State Update}}\\
		Let $\hat{{x}}_{k|k,0}=\hat{{x}}_{k|k-1}$ and obtain ${B}_{p}$ and ${B}_{r}$ by \eqref{bpbr}
		\While{$\frac{\left\|\hat{{x}}_{k|k,t}-\hat{{x}}_{k|k,t-1}\right\|}{\left\|\hat{{x}}_{k|k,t}\right\|}>\varepsilon$ or $t=1$}\\
		$\hat{{x}}_{k|k,t}=\hat{{x}}_{k|k-1}+\tilde{{K}}_{k,t}({y}_k-{H}_k\hat{{x}}_{k|k-1})$ \Comment{$t$ starts from 1}\\
		$\tilde{{K}}_{k,t}=\tilde{{P}}_{ k|k-1}{H}_k^{T}({H}_k\tilde{{P}}_{ k|k-1}{H}_k^{T}+{R}_{k})^{-1}$\\
		$\tilde{P}_{ k|k-1}={B}_{p}\tilde{{M}}_{p}^{-1}{{B}}_{p}^{T}$\\
		$ \tilde{{M}}_{p}=\mathrm{diag}({G}_{\boldsymbol{\sigma}_p}({e}_{p,k}))$ \Comment{$\boldsymbol{\sigma}_p=[\sigma_d^{T},\sigma_{s}^{T}]^{T}$}\\
		${e}_{p,k}={B}_{p}^{-1}\hat{{x}}_{k|k-1}-{B}_{p}^{-1}\hat{{x}}_{k|k,t-1}$\\
		$t \leftarrow t+1$
		\EndWhile\\
		${P}_{k|k}=({I}-\tilde{{K}}_k {H}_k){{P}}_{k|k-1}({I}-\tilde{{K}}_k {H}_k)^{T}+\tilde{{K}}_k {R}_k\tilde{{K}}_k^{T}$		
	\end{algorithmic}
\end{algorithm}

\begin{remark}
	The Gaussian kernel function obtains its maximum value when the error is equal to zero, which implies that $\tilde{{P}}_{ k|k-1} \succeq {{P}}_{ k|k-1}$. In MKCEKF-DOB, we apply ${\sigma}_s \to \infty$, which indicates that only the submatrix of ${{P}}_{ k|k-1}$ associating with the disturbance is inflated where the inflation level is determined by ${e}_{d,k}$ and ${\sigma}_d$. This mechanism can be understood as applying an ``adaptive'' process covariance matrix so that it matches the practical process disturbance error under the MKCL criterion. Benefiting from the fixed-point iteration in Line 8-15 of Algorithm \ref{mkcekfdob}, MKECKF-DOB alleviates the bias-variance dilemma compared with the conventional EKF-DOB.
\end{remark}
\begin{remark}
	The fixed-point iteration usually converges after 2-3 iterations, which indicates that MKCEKF-DOB has moderate computation complexities compared with EKF-DOB.
\end{remark}
\section{Simulations and Experiments}
In this section, we perform simulations and experiments to validate the effectiveness of the proposed methods.
\subsection{Simulations}
\begin{figure}[htbp]
	\centering 	
	\subfigure[]{
		\begin{minipage}[t]{0.5\linewidth}
			\centering
			\includegraphics[width=1.0\columnwidth]{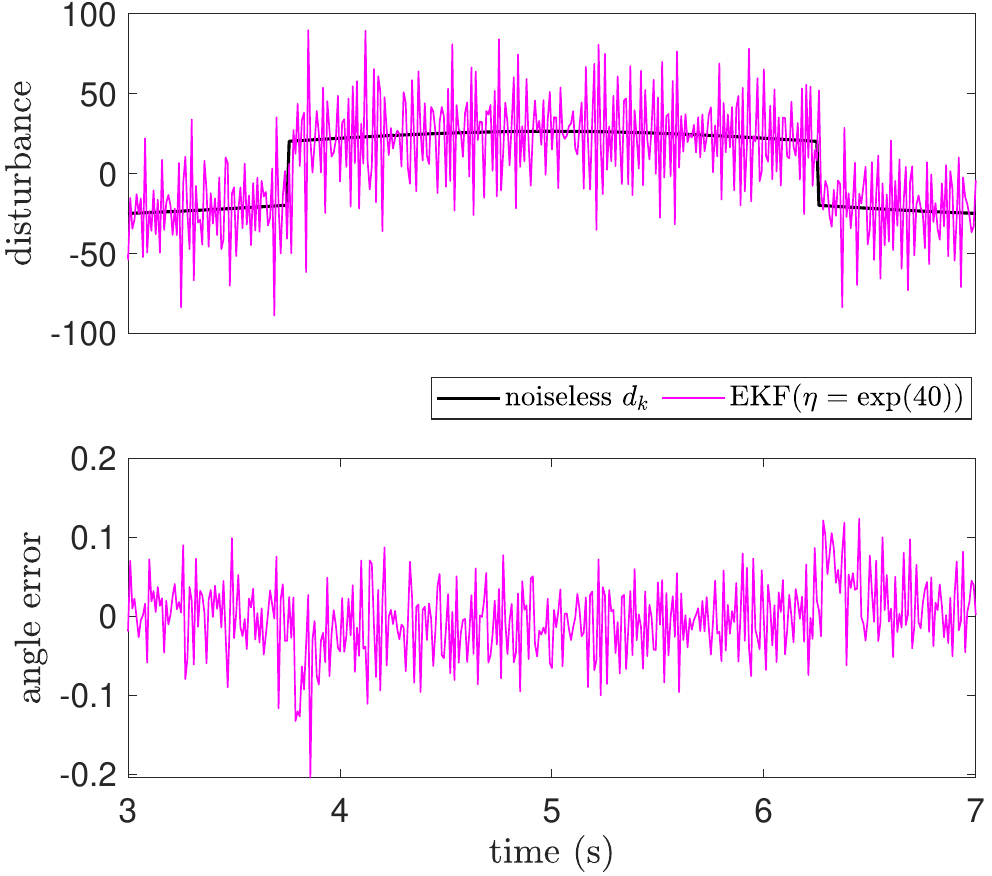}
			\label{pdobs0}
		\end{minipage}%
	}%
	\subfigure[]{
		\begin{minipage}[t]{0.5\linewidth}
			\centering
			\includegraphics[width=1.0\columnwidth]{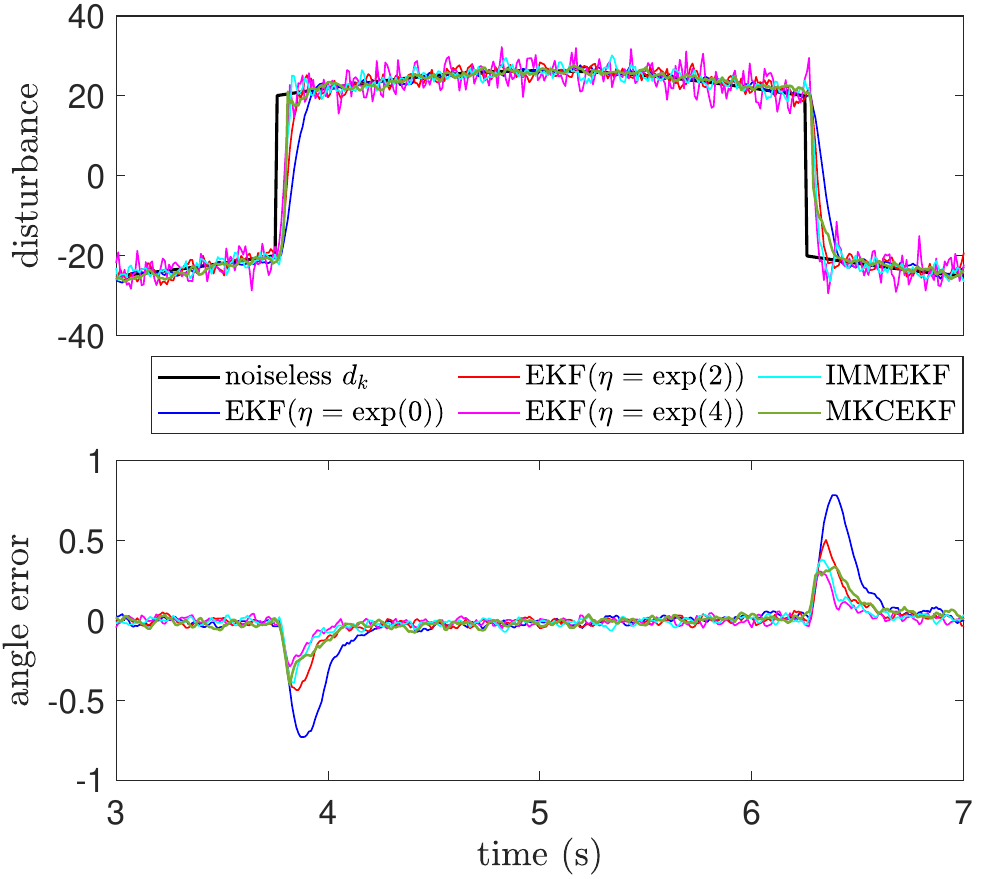}
			\label{pdobs1}
		\end{minipage}%
	}%
	\textcolor{black}{
		\caption{The disturbance estimation and tracking errors in EKF-DOB, IMMEKF-DOB, and MKCEKF-DOB. (a) Results with $\eta=\exp(40)$. (b) Comparisons of EKF-DOB, IMMEKF-DOB, and MKCEKF-DOB}
		\label{pdobs}}
\end{figure}

We consider a one-degree of freedom robotic manipulator tracking problem with unknown disturbance. The system dynamics is shown in \eqref{sysdyn}. The corresponding discrete time state-space model (with addictive noise) can be written as 
\begin{equation}
	\begin{aligned}
		x_{k}&=f(x_{k-1},u_{k-1})+w_{k-1} \\
		y_k&= g(x_k) + v_k
	\end{aligned}
\end{equation}
where $x_k=[d_k,\dot{\theta}_k, \theta_k]$ contains the disturbance, angular velocity, and angle. The process and observation functions have 
\begin{equation}
	\begin{aligned}
		&f(x_{k-1},u_{k-1})=\\
		&\begin{bmatrix}
			d_{k-1}\\
			\frac{\Delta T}{\mathrm{I}}\big(u_{k-1}+d_{k-1}+\frac{\mathrm{I}-\mathrm{b}\Delta T}{\Delta T}\dot{\theta}_{k-1}-\mathrm{k}\theta_{k-1}-\mathrm{m}g\sin(\theta_{k-1})\big) \\
			\dot{\theta}_{k-1} \Delta T + \theta_{k-1}
		\end{bmatrix}\\
		&g(x_k) =H_k x_k, ~H_k=[0,0,1]^{T}.
	\end{aligned}
\end{equation} 
\begin{table*}[htbp]
	\centering
	\caption{\textcolor{black}{Performance of Different Observers with Step-like Disturbance.}}
	\renewcommand{\arraystretch}{0.85}
	\begin{center}
		\scalebox{0.9}{
			\begin{tabular}{ccccccc}
				\hline
				\hline
				\multirow{2}{*}{Observer} & \multirow{2}{*}{\begin{tabular}[c]{@{}c@{}}RMSE\\ of $x_1$ \end{tabular}} & \multirow{2}{*}{\begin{tabular}[c]{@{}c@{}}RMSE\\ of ${x}_{2}$\end{tabular}} & \multirow{2}{*}{\begin{tabular}[c]{@{}c@{}}RMSE\\ of ${x}_{3}$\end{tabular}} & \multirow{2}{*}{\begin{tabular}[c]{@{}c@{}}RMSE\\ of $(\theta_d-\theta)$\end{tabular}} &
				\multirow{2}{*}{\begin{tabular}[c]{@{}c@{}}RMSE\\ of $(\dot\theta_d-\dot\theta)$\end{tabular}}& \multirow{2}{*}{\begin{tabular}[c]{@{}c@{}}time\\cost (s)\\ \end{tabular}} \\                                                               
				& & & & & &\\
				\hline
				{EKF-DOB $\eta=\exp(0)$}&6.887 $\pm$ 0.040 & 2.020 $\pm$ 0.029 & 0.0119 $\pm$ 0.0003&0.188 $\pm$ 0.0010& 2.132 $\pm$ 0.037 & 0.0025 $\pm$ 0.001 \\			
				\hline
				{EKF-DOB $\eta=\exp(1)$}&6.270 $\pm$ 0.050 & 1.548 $\pm$ 0.035 & 0.0100 $\pm$ 0.0003&0.134 $\pm$ 0.0008& 1.862 $\pm$ 0.041 & 0.0025 $\pm$ 0.001 \\
				\hline
				{EKF-DOB $\eta=\exp(2)$}&5.830 $\pm$ 0.061 & 1.272 $\pm$ 0.031 & 0.0092 $\pm$ 0.0002&0.098 $\pm$ 0.0007& 1.704 $\pm$ 0.040 & 0.0025 $\pm$ 0.001 \\
				\hline
				{EKF-DOB $\eta=\exp(3)$}&5.575 $\pm$ 0.081 & 1.142 $\pm$ 0.036 & 0.0090 $\pm$ 0.0002&0.074 $\pm$ 0.0006& 1.652 $\pm$ 0.043 & 0.0024 $\pm$ 0.001\\
				\hline
				{EKF-DOB $\eta=\exp(4)$}&5.618 $\pm$ 0.091 & 1.154 $\pm$ 0.029 & 0.0090 $\pm$ 0.0002&0.058 $\pm$ 0.0006& 1.740 $\pm$ 0.038 & 0.0025 $\pm$ 0.001\\
				\hline
				{EKF-DOB $\eta=\exp(40)$}&27.038 $\pm$ 0.793 & 4.077 $\pm$ 0.117 & 0.0100 $\pm$ 0.0002&0.048 $\pm$ 0.0010& 7.146 $\pm$ 0.209 & 0.0024 $\pm$ 0.001 \\
				\hline
				{IMMEKF-DOB}&5.574 $\pm$ 0.063 & 1.124 $\pm$ 0.029 & 0.0089 $\pm$ 0.0003&0.074 $\pm$ 0.0008& 1.633 $\pm$ 0.042 & 0.0069 $\pm$ 0.001 \\
				\hline
				{MKCEKF-DOB}&5.472 $\pm$ 0.071 & 1.030 $\pm$ 0.033 & 0.0086 $\pm$ 0.0002&0.077 $\pm$ 0.0026& 1.496 $\pm$ 0.047 & 0.0058 $\pm$ 0.001\\
				\hline
				\hline
		\end{tabular}}
		\label{gauDisturbnace}
	\end{center}
\end{table*}
In simulation, the desired angle follows $\theta_{d,k}=10\sin(0.4\pi k T)$ where $T=0.01$ and $k$ is the time index. The system parameters have $\mathrm{I}=0.1$, $\mathrm{m}=0.1$, $\mathrm{k}=0.1$, $\mathrm{b}=1$, $l=0.2$, $g=9.81$. The disturbance is considered as Coulomb and viscous friction:
\begin{equation}
	{d}_{k}=20\operatorname{sign}(\dot{\theta})+0.5\dot{\theta}+{w}_{d,k}
	\label{stepdist}
\end{equation}
where ${w}_{d,k} \sim  \mathcal{N}(0,0.25)$. We apply the augmented PD controller (see \eqref{controller}), including a feedforward term, a feedback term, and a disturbance compensation term. In different estimators, the process covariance follows the form
$Q=\begin{bmatrix}
	\eta Q_{d}&0 \\
	0& Q_{s}
\end{bmatrix}
$ where $Q_{d}=0.25$ and $\eta \ge 1$ is a scalar. In EKF-DOB, we vary $\eta$ from $\exp(0)$ to $\exp(4)$ to investigate the bias-variance effect with different $\eta$. The results are shown in Fig. \ref{pdobs}. We observe that an inherent disturbance tracking speed and tracking variance exist. Specifically, as $\eta = \exp(40)$, the estimator becomes unbiased, at the cost of the maximum disturbance uncertainty, which is aligned with \cite{bb29}.

We then compare EKF-DOB with IMMEKF-DOB and MKCEKF-DOB. In IMMEKF-DOB, two different disturbance noise covariance, $Q_{d1}=\exp(0)Q_{d}$ and $Q_{d2}=\exp(4)Q_{d}$, are used with Markov transition probability matrix 
$P=\begin{bmatrix}
	0.95&0.05\\
	0.3&0.7
\end{bmatrix}$. In MKCEKF-DOB, we apply $\exp(0)Q_{d}$ with kennel bandwidth $\sigma_d=1.5$. The comparison is shown in Fig. \ref{pdobs1}, indicating that IMMEKF-DOB and MKCEKF-DOB outperform the best EKF-DOB. The performance of different algorithms, evaluated over 100 independent Monte Carlo runs, is summarized in Table \ref{gauDisturbnace}. We find that the disturbance estimation accuracy of IMMEKF-DOB and MKCEKF-DOB improves by 60.6\% and 59.0\%, respectively, compared with EKF-DOB ($\eta=\exp(0)$), demonstrating significant enhancement using the same process and measurement covariance matrices.

\begin{figure}[htbp]
	\centering 	
	\subfigure[]{
		\begin{minipage}[t]{0.45\linewidth}
			\centering
			\includegraphics[width=1.0\columnwidth]{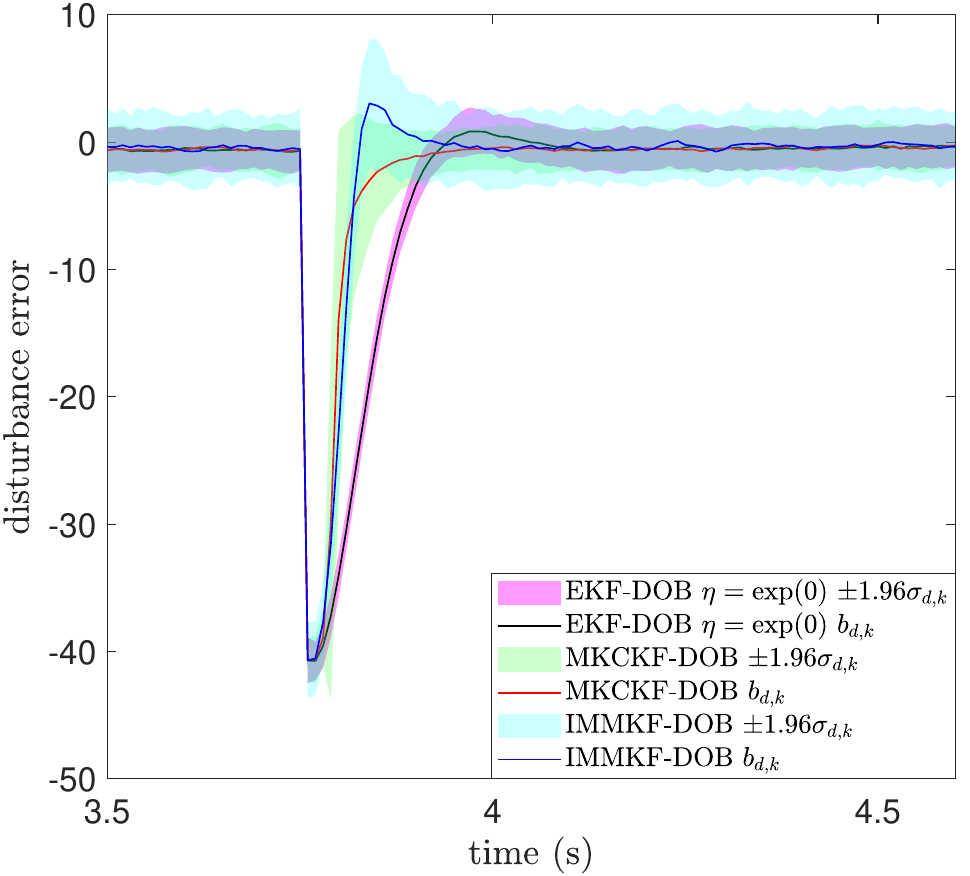}
			\label{dbiasvar}
		\end{minipage}%
	}%
	\subfigure[]{
		\begin{minipage}[t]{0.45\linewidth}
			\centering
			\includegraphics[width=1.0\columnwidth]{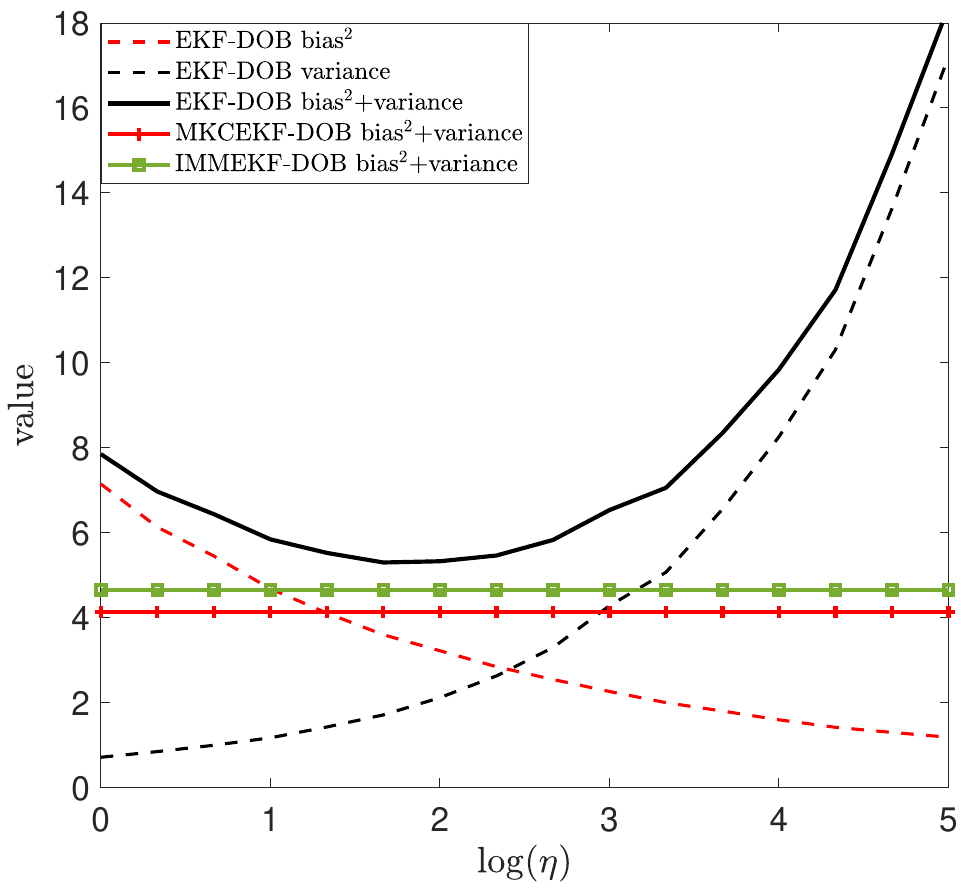}
			\label{tradeoff}
		\end{minipage}%
	}%
	\caption{The visualization of bias-variance trade-off in EKF-DOB, IMMEKF-DOB, and MKCEKF-DOB. (a) Bias-variance effects of different algorithms in time series (95\% confidence interval). (b) Bias-variance effects of EKF-DOB with different $\eta$ and corresponding results of IMMEKF-DOB and MKCEKF-DOB. Note that IMMEKF-DOB and MKCEKF-DOB are not a function of $\eta$ and hence are visualized as flat lines.}
	\label{bias_variance_tradeoff}
\end{figure}

To demonstrate the bias-variance trade-off of different algorithms, we compare the bias and standard deviation of the disturbance estimation at each time step with different $\eta$ and different algorithms. Note that the average disturbance bias is obtained by $b_{d,k}= \frac{1}{K}\sum_{i=1}^{K} (d_k -\hat{d}_k)$ and the standard deviation is obtained by $\sigma_{d,k}=\sqrt{\frac{1}{K}{\sum_{i=1}^{K}(d_k -\hat{d}_k-b_{d,k})^{2}}}$ where $K$ is the Monte Carlo counts. We calculate the average square of bias as $\bar{b}_{d}^{2}=\frac{1}{m_2-m_1+1}\sum_{k=m_1}^{m_2} b_{d,k}^{2} $ and average variance as $\bar{\sigma}^2=\frac{1}{m_2-m_1+1}\sum_{k=m_1}^{m_2} \sigma_{k}^2$ within the time interval $k=[300,450]$ (since this region contains both the slow-varying disturbance and disturbance jump). The corresponding results are shown in Figs. \ref{dbiasvar} and \ref{tradeoff}, where one can observe that IMMEKF-DOB and MKCEKF-DOB are better than EKF-DOB with any $\eta$. 

\subsection{Angle Tracking Experiments}
\label{angle tracking}
\begin{figure}[htbp]
	\centering
	\subfigure[]{
		\begin{minipage}[t]{0.35\linewidth}
			\centering
			\includegraphics[width=0.85\columnwidth]{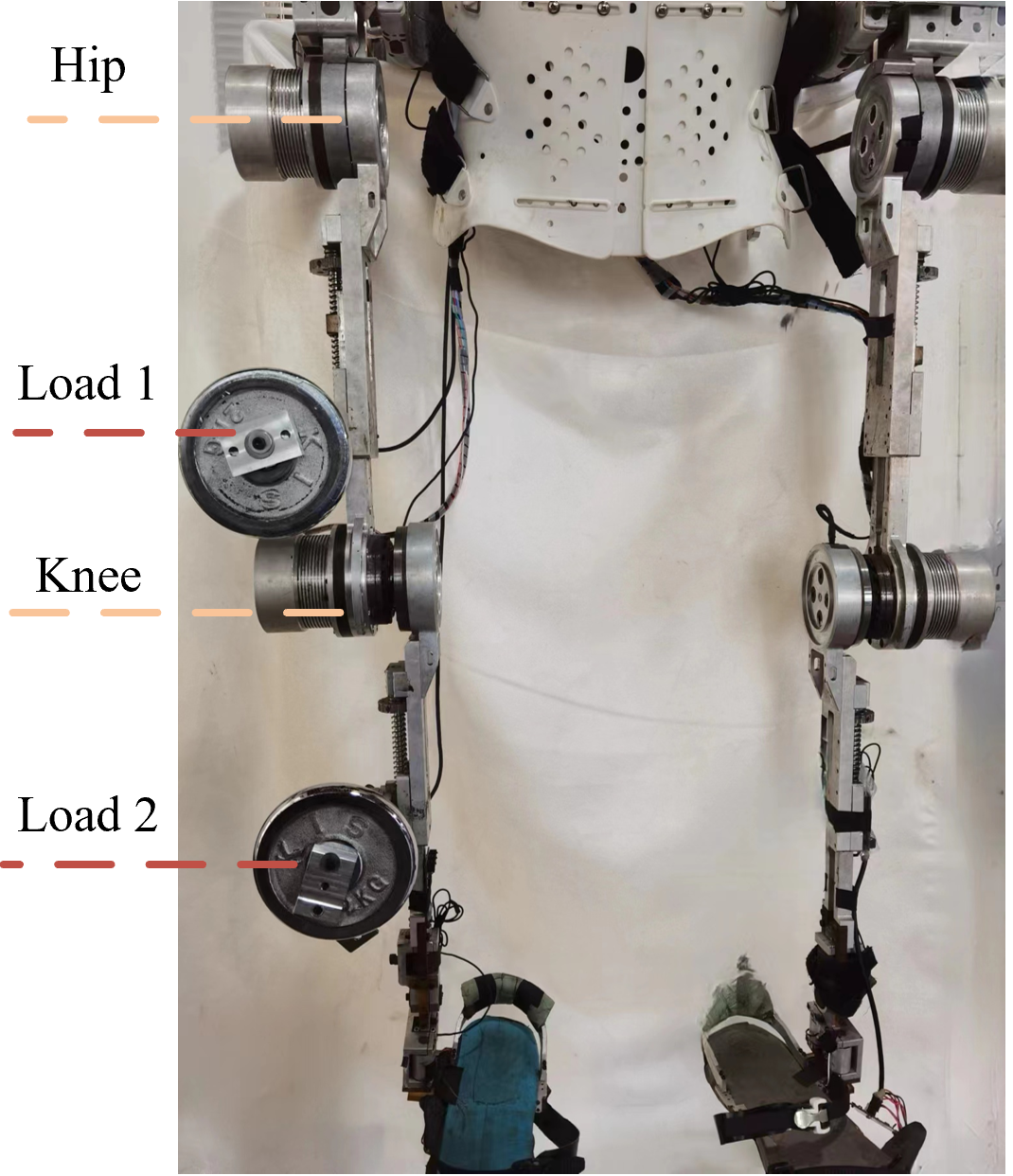}
			\label{exp_setup}
		\end{minipage}%
	}%
	\subfigure[]{
		\begin{minipage}[t]{0.35\linewidth}
			\centering
			\includegraphics[width=1.0\columnwidth]{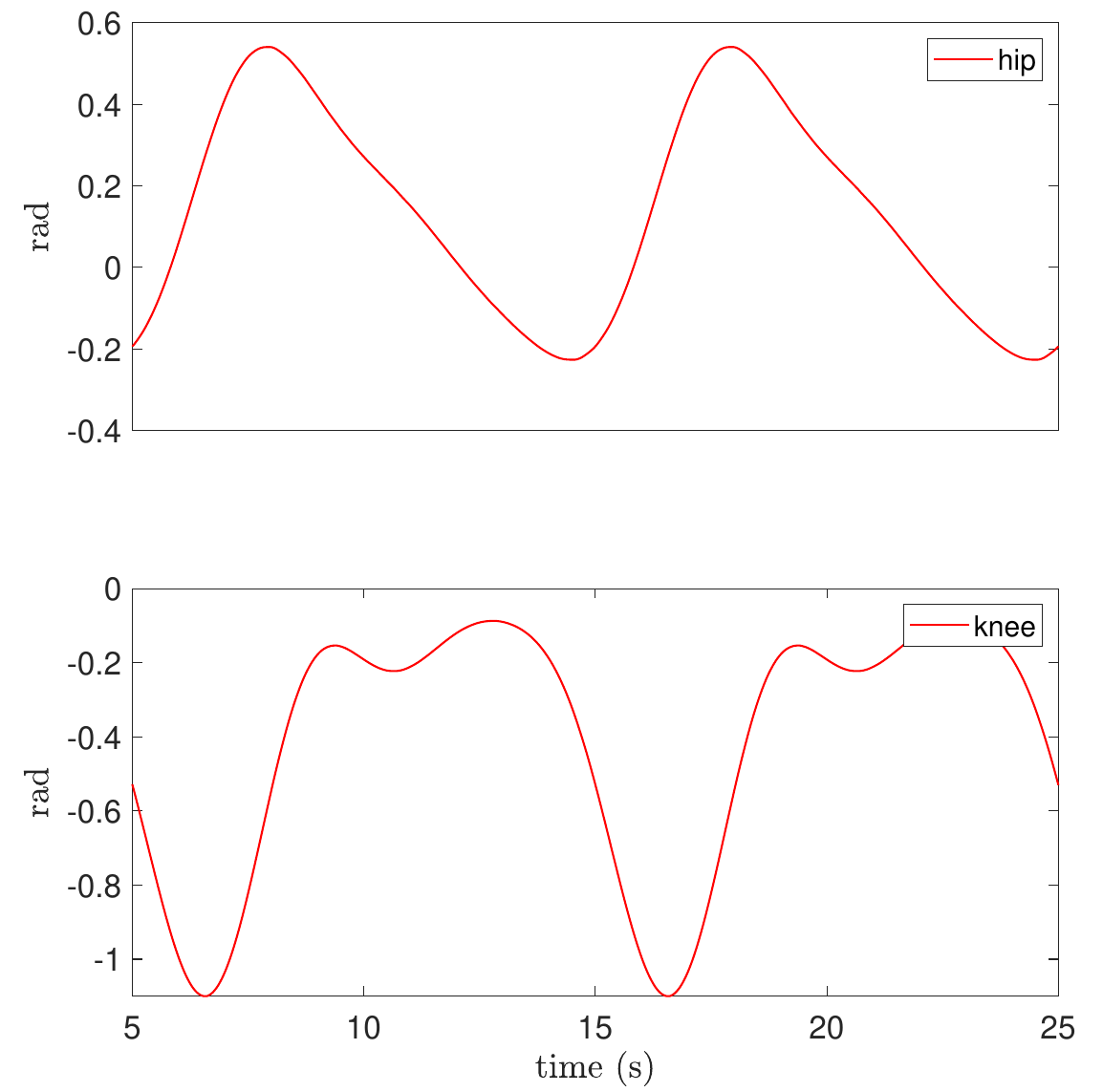}
		\end{minipage}%
		\label{pos_f10_cmd1}
	}%
	\caption{The experimental setup and commanded angle. (a) The experimental setup. (b) The commanded hip and knee angle with $f=0.1$ Hz.}	
	\label{exp}
\end{figure}

We conduct angle-tracking experiments on exoskeletons using MATLAB/Simulink Real-Time at a control frequency of 1 kHz. The communication between the computer and the exoskeleton is established via the EtherCAT protocol. The host computer is configured with Windows 7, an Intel i5-4460 quad-core CPU, 4 GB of RAM, and MATLAB 2017b. The experimental setup is shown in Fig. \ref{exp_setup}, where two unknown loads are attached to the left thigh and left shank to simulate the influence of a human leg when conducting physical therapy using exoskeletons \cite{bb27}. Since the exoskeleton is symmetrical, we focus on the left side of the exoskeleton in experiments. The commanded hip and knee angle at frequency $f=0.1$ Hz is shown in Fig. \ref{pos_f10_cmd1}. 

\begin{figure}[htbp]
	\centering
	\subfigure[]{
		\begin{minipage}[t]{0.39\linewidth}
			\centering
			\includegraphics[width=1.0\columnwidth]{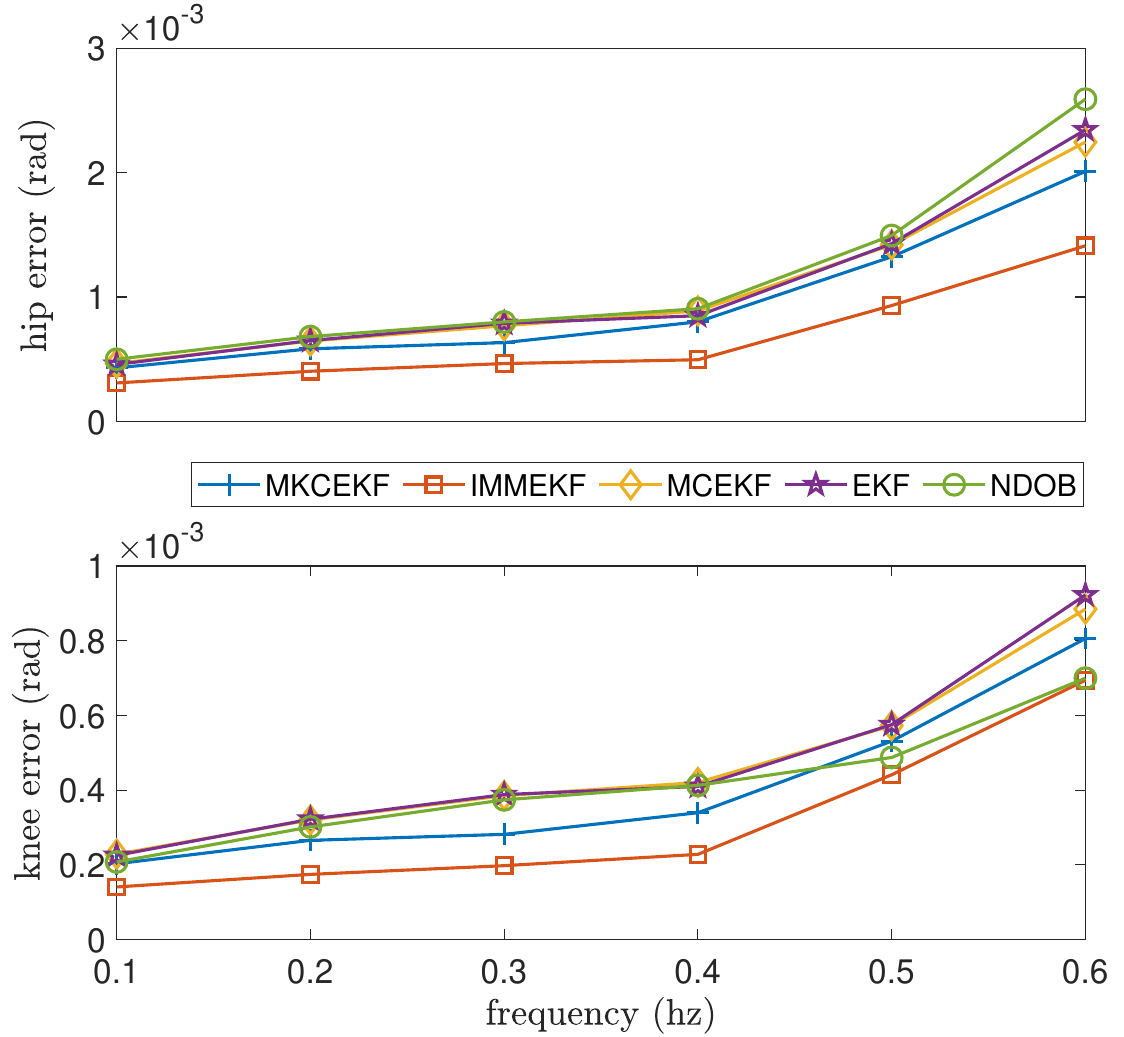}
		\end{minipage}%
		\label{pose}
	}%
	\subfigure[]{
		\begin{minipage}[t]{0.39\linewidth}
			\centering
			\includegraphics[width=1.0\columnwidth]{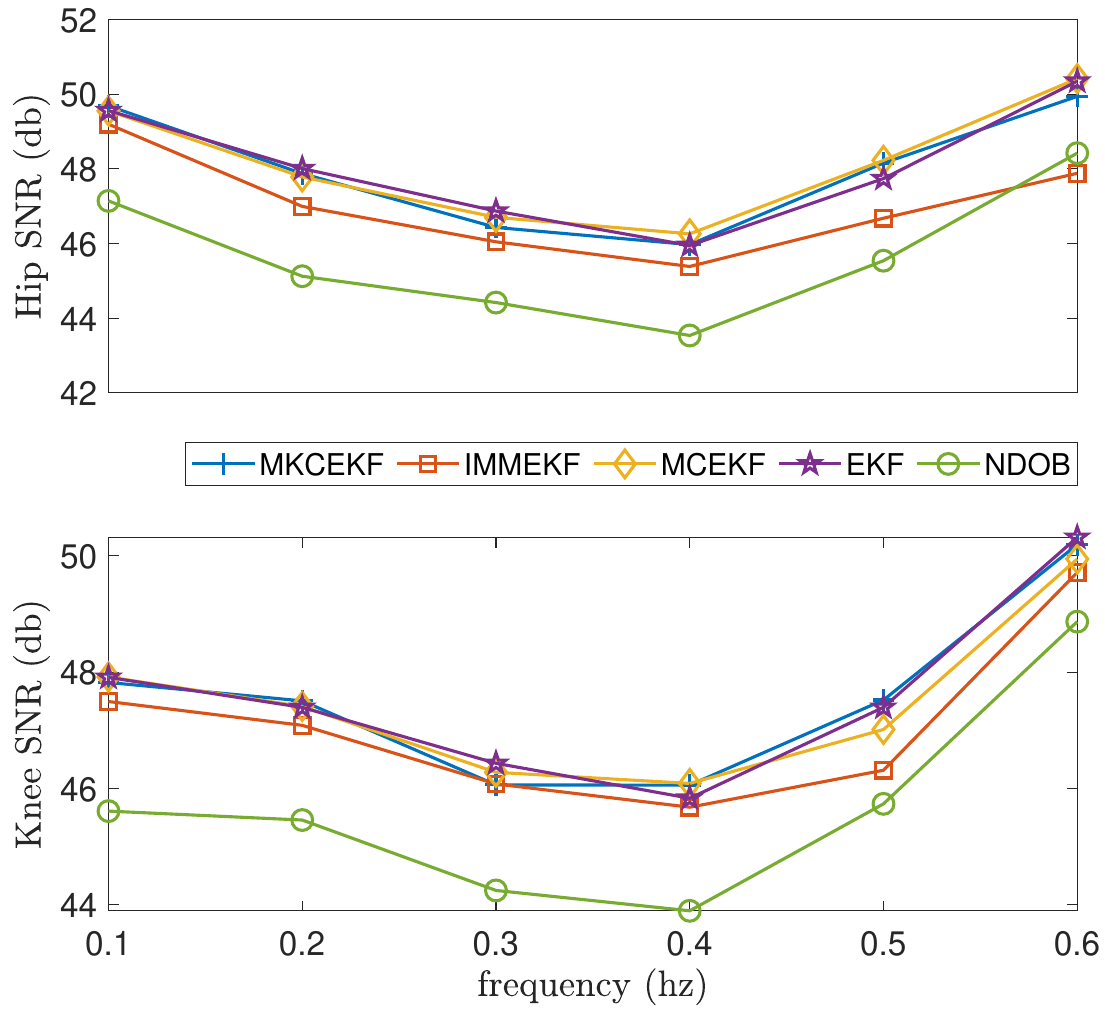}
		\end{minipage}%
		\label{snr}
	}%
	\caption{Tracking errors and  SNRs of different observers. (a) The hip and knee angle tracking errors. (b) The hip and knee SNRs.}	
\end{figure}

We employ the augmented PD controller accompanied by the disturbance observer in experiments. The controller is the same with \eqref{controller} using $K_p=\operatorname{diag}{[5000,5000]^{T}}$ and $K_d=\operatorname{diag}{[100,100]^{T}}$. We employ the EKF-DOB as a baseline to jointly estimate both the disturbance and the state. Additionally, we implement the MKCEKF-DOB, IMMEKF-DOB, and MCEKF-DOB (see \cite{bb28}) for comparisons. To ensure a fair comparison, we apply identical covariance parameters for the EKF-DOB, MKCEKF-DOB, and MCEKF-DOB, and use $Q_{d1}=Q_{d}$ and $Q_{d2}=20Q_{d}$ with $P=\begin{bmatrix}
	0.99&0.01\\
	0.01&0.99
\end{bmatrix}$ for IMMEKF-DOB. We sweep gait frequencies from $0.1$ Hz to $0.6$ Hz at an increment of $0.1$ Hz. The hip and knee root-mean-square-error of different disturbance observers are summarized in Tables \ref{hiprmse} and \ref{kneermse}. The corresponding visualizativisualizationon is shown in Fig. \ref{pose}. One can observe that the tracking errors of IMMEKF-DOB and MKCEKF-DOB decrease 37.99\% and 12.17\%, respectively, compared with EKF-DOB, by calculating the summed hip and knee errors and taking the average across the different frequencies. 

\begin{table}[h]
	\centering
	\caption{Hip Joint RMSEs (unit: mrad) of Different Observers Under Different Frequencies.}
	\scalebox{0.78}{
		\begin{tabular}{ccccccc}
			\hline
			\hline
			$f$ (Hz) & No DOB & MKCEKF  & IMMEKF & MCEKF  & EKF   & NDOB   \\
			\hline
			0.1 &1.6570    &0.4307    &0.3094    &0.4665    &0.4582    &0.4997 \\
			0.2 &1.5675    &0.5838    &0.4035    &0.6466    &0.6502    &0.6818 \\
			0.3 &1.6557    &0.6333    &0.4653    &0.7686    &0.7887    &0.7997 \\
			0.4 &1.7541    &0.8004    &0.4959    &0.8886    &0.8496    &0.9060 \\
			0.5 &2.4105    &1.3226    &0.9312    &1.4188    &1.4283    &1.4939 \\
			0.6 &3.5292    &2.0095    &1.4115    &2.2450    &2.3430    &2.5891 \\
			\hline
			\hline
	\end{tabular}}
	\label{hiprmse}
\end{table}

\begin{table}[h]
	\centering
	\caption{Knee Joint RMSEs (unit: mrad) of Different Observers Under Different Frequencies.}
	\scalebox{0.78}{
		\begin{tabular}{lllllll}
			\hline
			\hline
			$f$ (Hz) & No DOB & MKCEKF  & IMMEKF & MCEKF  & EKF   & NDOB   \\
			\hline
			0.1 &1.6539    &0.2026    &0.1405    &0.2284    &0.2251    &0.2075 \\
			0.2 &1.7875    &0.2653    &0.1742    &0.3199    &0.3219    &0.3013 \\
			0.3 & 1.8453    &0.2814    &0.1975    &0.3851    &0.3879    &0.3738\\
			0.4 & 1.8910    &0.3393    &0.2278    &0.4203    &0.4097    &0.4126\\
			0.5 & 1.9472    &0.5310    &0.4409    &0.5724    &0.5752    &0.4872\\
			0.6 & 1.9846    &0.8057    &0.6941    &0.8846    &0.9225    &0.6996\\
			\hline
			\hline
	\end{tabular}}
	\label{kneermse}
\end{table}

Since the signal-to-noise ratio (SNR) of the command signal reflects how much noise is involved in the command torque (i.e., $u_k$), we use this metric to evaluate the performance of different observers by investigating the SNR of different controller outputs. A zero-phase 12th-order Butterworth filter (cutoff frequency: 0.2) is applied to the torque command to separate the desired torque from the noise. The smoothed output is taken as the desired torque. In this way, we obtain the SNR performance of different controllers (observers) in Fig. \ref{snr}. We observe that NDOB has the lowest SNR while exhibiting almost the highest RMSEs (see Fig. \ref{pose}) compared to the other observers, indicating that its performance is inferior to the KF-based estimators. Additionally, the SNR of MKCEKF-DOB is similar to that of EKF-DOB and higher than that of IMMEKF-DOB, suggesting that the superior performance of IMMEKF-DOB in Fig. \ref{pose} partially results from its aggressive torque outputs.
\subsection{Dynamic Disturbance Experiments}
\begin{figure}[htbp]
	\centering
	\subfigure[Hand push experiment]{
		\begin{minipage}[t]{0.4\linewidth}
			\centering
			\includegraphics[width=0.55\columnwidth]{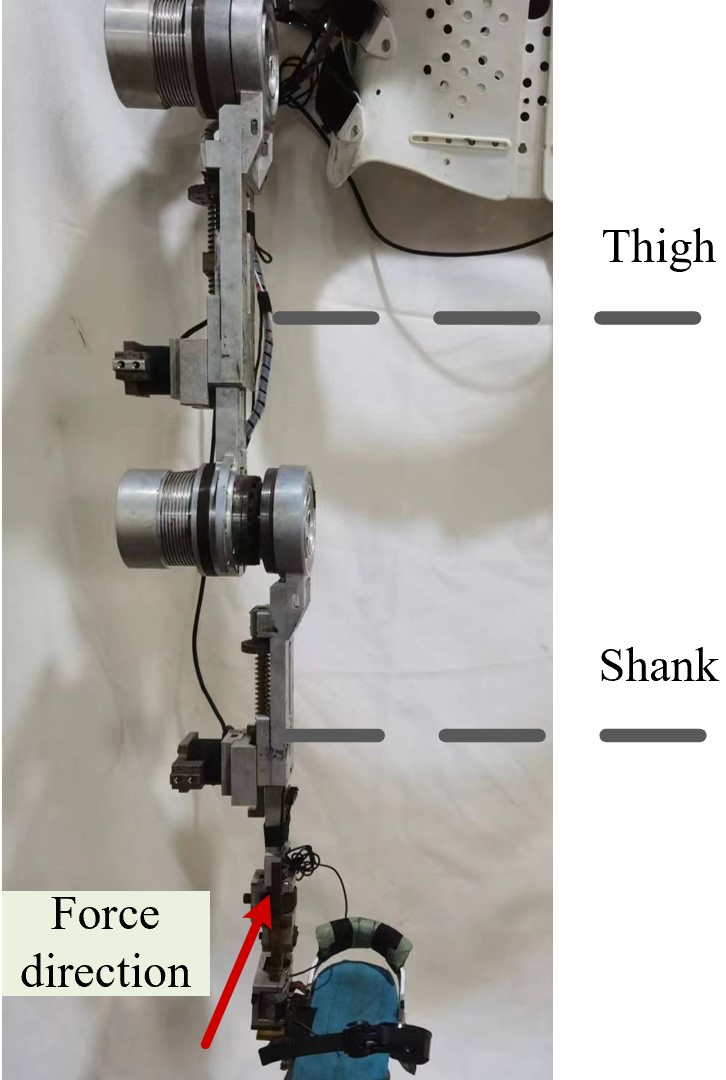}
			\label{exp_push}
		\end{minipage}%
	}%
	\subfigure[Elastic band experiment]{
		\begin{minipage}[t]{0.4\linewidth}
			\centering
			\includegraphics[width=0.6\columnwidth]{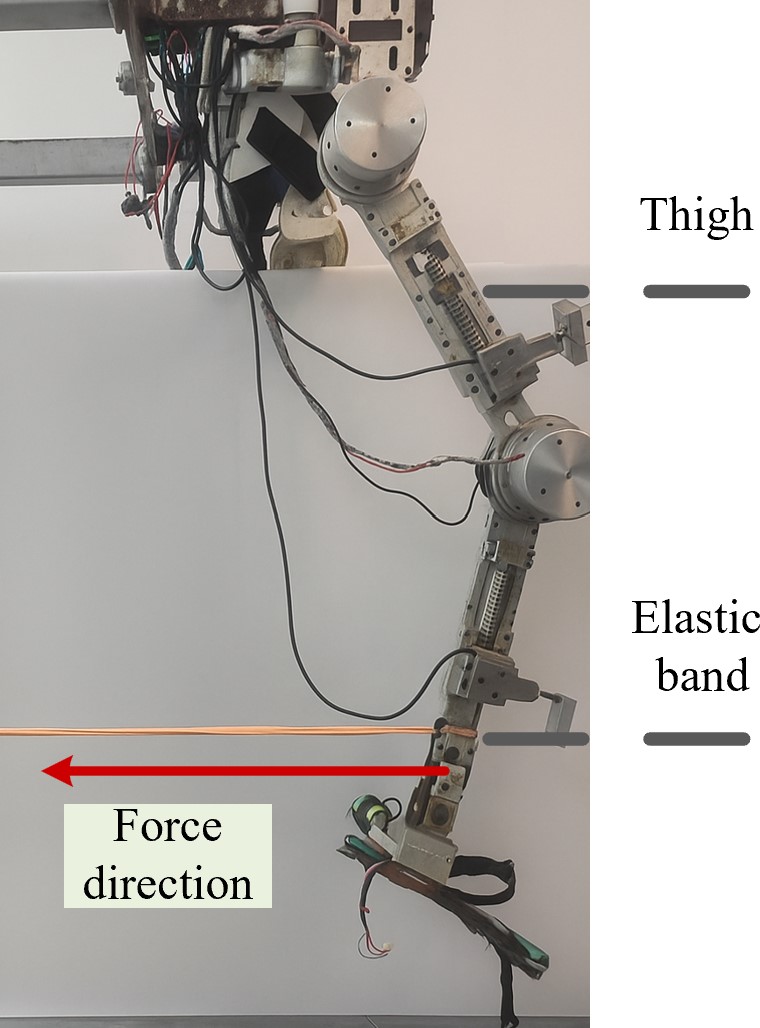}
		\end{minipage}%
		\label{pos_f10_cmd}
	}%
	\textcolor{black}{
		\caption{Dynamic disturbance experiments. (a) Hand push experiment: An unknown external force is applied to the exoskeleton by the subject's hand during the heel-strike phase of the gait cycle to simulate fast-varying disturbances. (b) Elastic band experiment: An elastic band is attached to the shank of the exoskeleton, with the other end fixed to a stationary frame. The band is stretched to varying lengths in sync with the gait trajectory, simulating different levels of disturbance. According to the product specifications, the maximum force exerted by the elastic band is less than 5 kg.}	\label{exp1}}
\end{figure}

To evaluate the performance of the proposed methods under dynamic disturbance scenarios, we conduct experiments involving external disturbances. In the first experiment, the subject occasionally pushes the shank of the exoskeleton during the heel strike phase of the gait cycle to simulate an impulsive-like disturbance. In the second experiment, a cyclic and repeatable disturbance is generated by an elastic band attached to the shank of the exoskeleton, with the other end fixed to a stationary frame. The detailed experimental setup is illustrated in Fig.~\ref{exp1}. In both experiments, the disturbance observer parameters are the same as those used in the previous section, and the gait frequency is maintained at 0.3 Hz.

\begin{figure}[htbp]
	\centering
	\includegraphics[width=0.4\linewidth]{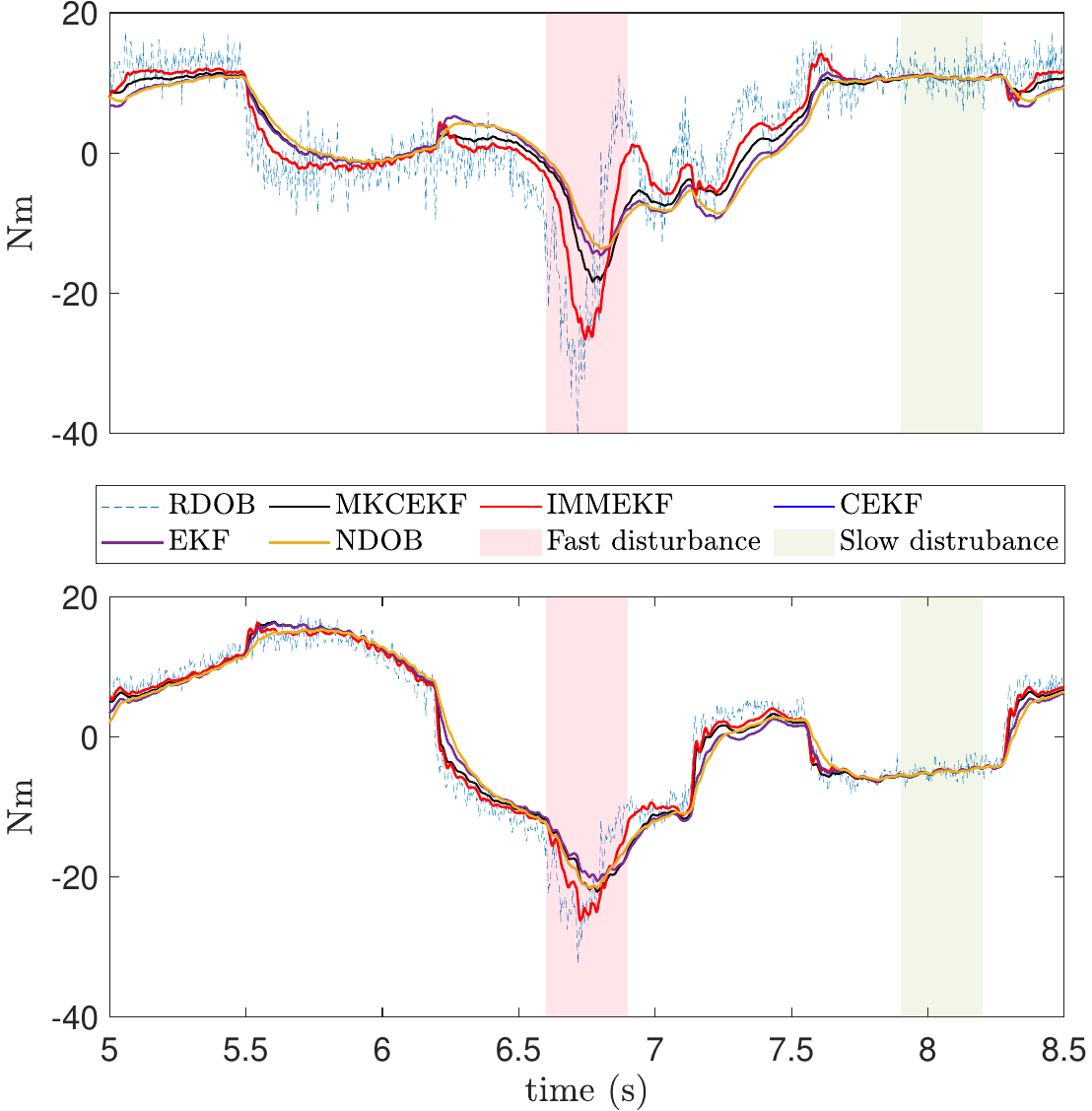}
	\textcolor{black}{
		\caption{Performance comparison of different disturbance observers in the hand push experiment. The pink area indicates the region of hand-applied pushes, representing fast-varying disturbances. The green area marks the region of slow-varying disturbances. It can be observed that IMMKF and MKCEKF exhibit fast disturbance tracking in the fast-varying region, while maintaining similar smoothness to the EKF in the slow-varying region.}    \label{handpush}
	}
\end{figure}

\textcolor{black}{
	The comparison of different observers in the hand push experiment is shown in Fig. \ref{handpush}. The RDOB, derived from \eqref{sysdyn}, provides an unbiased but noisy estimate of the external disturbance. Hence, it can be seen as a noisy version of the ground truth disturbance. We observe that the proposed observers, MKCEKF and IMMKF, achieve similar smoothness to the EKF in regions of slow-varying disturbances (mainly due to viscous friction), while exhibiting significantly faster tracking in regions with fast-varying disturbances. These results confirm that the proposed approaches are well-suited for scenarios involving both fast- and slow-varying disturbances.}
\begin{figure}[htbp]
	\centering
	\includegraphics[width=0.55\linewidth]{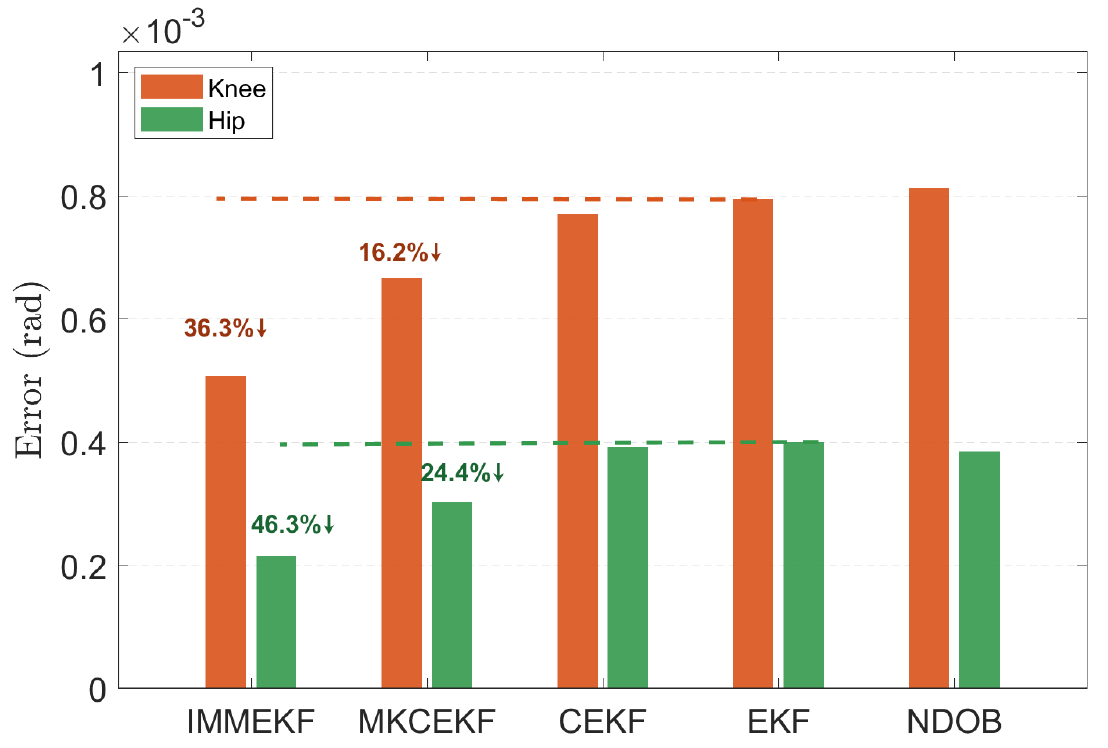}
	\textcolor{black}{
		\caption{ Tracking error of different disturbance observers in the elastic band experiment.}    \label{ElasticBand}}
\end{figure}

The tracking performance of different observers in the elastic band experiment is presented in Fig.~\ref{ElasticBand}. The RMSE of the tracking error for the EKF is $0.795 \times 10^{-3}$ rad for the hip joint and $0.400  \times 10^{-3}$ rad for the knee joint. In comparison, the IMMEKF achieves significantly lower errors of $0.507 \times 10^{-3}$ rad (hip) and $0.215 \times 10^{-3}$ rad (knee), while the MKCEKF yields $0.666 \times 10^{-3}$ rad (hip) and $0.302 \times 10^{-3}$ rad (knee). These results correspond to reductions of $36.3\%$ and $16.2\%$ in hip joint error, and $46.3\%$ and $24.4\%$ in knee joint error, for the IMMEKF and MKCEKF, respectively, compared to the EKF, showing the superiority of the proposed remedies. We further conduct experiments to investigate the influence of observer parameters and the performance of the proposed methods on six-degree-of-freedom manipulators, see \textcolor{black}{Section \ref{abc}} for details. All these experiments or simulations verified the utility and the effectiveness of the proposed methods. 
\section{Conclusion}
In this paper, we systematically investigate the bias-variance effects in the EKF-DOB for angle tracking of exoskeletons. To achieve timely and smooth disturbance estimation, we propose two solutions: MKCEKF-DOB and IMMEKF-DOB. The superiority of these methods is attributed to their ``adaptive" or ``switched" disturbance covariance and is validated through extensive simulations and experiments. Notably, while MKCEKF-DOB and IMMKF-DOB demonstrate improved performance compared to existing approaches, MKCEKF-DOB requires tuning of the kernel bandwidths, and IMMKF-DOB necessitates empirical design of the Markov transition probability matrix. In future work, we aim to develop adaptive mechanisms to streamline these time-consuming procedures.
\section{Appendix I}
\subsection{Proof of Lemma \ref{lemma2}}
\label{appen1}
\begin{proof}
	For system \eqref{closesys}, we construct the following Lyapunov function:
	\begin{equation}
		V(e, \dot{e})=\frac{1}{2} \dot{e}^T M(\theta) \dot{e}+\frac{1}{2} e^T K_p e+\epsilon e^T M(\theta) \dot{e}
	\end{equation}
	which is positive definite for sufficient small $\epsilon$ since $M(\theta) \succ 0$ and $K_p \succ 0$.  
	According to \cite{bb20}, $\dot{M}-2C(\theta,\dot{\theta})$ is a skew-symmetric matrix. It follows that 
	\begin{equation}
		\begin{aligned}
			\dot{V}&=\dot{e}^T M \ddot{e}+\frac{1}{2} \dot{e}^T \dot{M} \dot{e}+\dot{e}^T K_p e+\epsilon \dot{e}^T M \dot{e}+\epsilon e^T(M \ddot{e}+\dot{M} \dot{e})\\
			&=\underbrace{-\dot{e}^T\left(K_d-\epsilon M\right) \dot{e}-\epsilon e^T K_p e+\epsilon e^T(-K_d+\frac{1}{2} \dot{M}) \dot{e}}_{P_1}\\
			&\underbrace{-(\dot{e}^T+\epsilon e^T)l_e}_{P_2}.
		\end{aligned}
	\end{equation}
	Choosing sufficiently small positive $\epsilon$ guarantees that $P_1$ is negative definite~\cite{bb20}. Then, by applying Young's inequality, one has
	\begin{equation}
		|P_2| \le \frac{1}{2} \dot{e}^T K_d \dot{e}+\frac{1}{2} l_e^T K_d^{-1} l_e + \frac{\epsilon}{2} {e}^T K_p {e}+\frac{\epsilon}{2} l_e^T K_p^{-1} l_e.
	\end{equation}
	It follows that
	\begin{equation}
		\begin{aligned}
			\dot{V} &\le P_1 + \frac{1}{2} \dot{e}^T K_d \dot{e}+\frac{1}{2} d_e^T K_d^{-1} d_e + \frac{\epsilon}{2} {e}^T K_p {e}+\frac{\epsilon}{2} d_e^T K_p^{-1} d_e \\
			&= \underbrace{-\dot{e}^T(\frac{1}{2}K_d-\epsilon M) \dot{e}-\frac{1}{2} \epsilon e^T K_p e+\epsilon e^T(-K_d+\frac{1}{2} \dot{M}) \dot{e}}_{P_1^{\prime}}\\
			&+ \underbrace{\frac{1}{2} l_{e}^T K_d^{-1} l_{e}+\frac{\epsilon}{2} l_{e}^T K_p^{-1} l_{e}}_{P_2^{\prime}}.
		\end{aligned}
		\label{vdot}
	\end{equation}
	It holds that \begin{equation}
		P_1^{\prime} \preceq -\alpha_1 {e}^{T}{e} -\alpha_2 \dot{e}^{T}\dot{e}
		\label{expconv}
	\end{equation} for sufficiently small $\epsilon>0$, $\alpha_1>0$, and $\alpha_2 >0$~\cite{bb20}. Note that $P_2^{\prime}$ is bounded since $d_e$ is bounded. Hence, the system is globally uniformly ultimately bounded. 
\end{proof}
\subsection{Proof of Corollary \ref{corollary1}}
\label{appen2}
\textcolor{black}{
	\begin{proof}
		From \eqref{vdot} and \eqref{expconv}, we obtain:
		\[
		\dot{V} \leq -\alpha_1 e^{T}e -\alpha_2 \dot{e}^{T}\dot{e} + \frac{\bar{l}_e^2}{2} \left( \frac{1}{\lambda_{\min}(K_d)} + \frac{\epsilon}{\lambda_{\min}(K_p)} \right).
		\]
		Define the positive definite quadratic form:
		\[
		W(e, \dot{e}) = \alpha_1 \|e\|^{2} + \alpha_2 \|\dot{e}\|^{2}.
		\]
		When $W(e, \dot{e}) > \dfrac{\bar{l}_e^2}{2} \left( \dfrac{1}{\lambda_{\min}(K_d)} + \dfrac{\epsilon}{\lambda_{\min}(K_p)} \right)$, we have:
		\[
		\dot{V} < -W + W = 0.
		\]
		Since $V$ is radially unbounded and $\dot{V} < 0$ outside the ellipsoid $\mathcal{B}$, all trajectories converge to and remain in:
		\[
		\mathcal{B} = \left\{ (e, \dot{e}) : W(e, \dot{e}) \leq \dfrac{\bar{l}_e^2}{2} \left( \dfrac{1}{\lambda_{\min}(K_d)} + \dfrac{\epsilon}{\lambda_{\min}(K_p)} \right) \right\}.
		\]
\end{proof}}
\section{Appendix II}
	\label{abc}
	\subsection{Exoskeleton Model and System Identification}
\label{section1}
\begin{figure}[htbp]
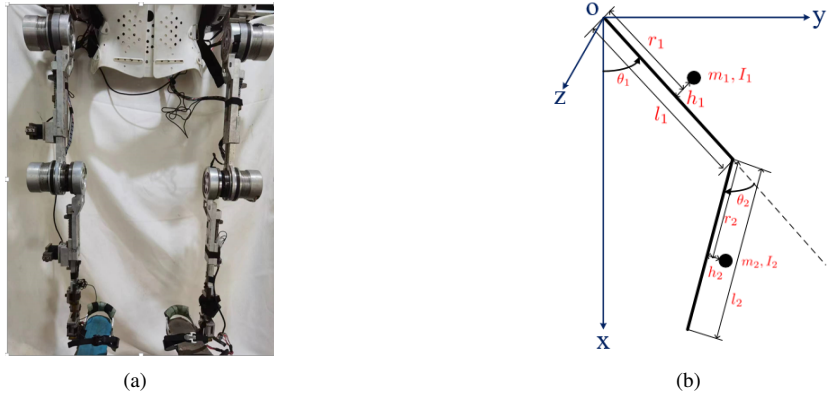

	\centering
	\subfigure[]{\includegraphics[width=1.4in]{EXO1.png}%
		\label{exo1}}
	\hfil
	\subfigure[]{\includegraphics[width=1.5in]{exo_link.png}%
		\label{exo_link1}}
	\caption{The exoskeleton and schematic diagram of two-link robotic leg model.}		

\end{figure}

The exoskeleton is shown in Fig. \ref{exo1}. In our applications, the exoskeleton's waist is fixed to a frame and hence each leg of the exoskeleton can be modeled as a two-link robotic leg. The corresponding schematic diagram is shown in Fig. \ref{exo_link1} where $m$ is the mass, $I$ is the inertia, $l$ is the link length, $r$ is the distance from joints to the center of mass along the direction of the link, $h$ is the distance from the center of mass to the link, and $\theta$ is the rotation angle. The subscript $1,2$ represents the corresponding values for link 1 and link 2. 

The locations of the center of mass for link 1 [$(x_1,y_1)$] and link 2 [$(x_2,y_2)$] can be written as
\begin{equation}
	\begin{aligned}
		x_1 &= r_1 c_1 -h_1 s_1\\
		y_1 &= r_1 s_1 + h_1 c_1\\
		x_2 &= l_1 c_1 + r_2 c_{12} - h_2 s_{12}\\
		y_2 &= l_1 s_1 + r_2 s_{12} + h_2 c_{12}
	\end{aligned}
\end{equation}
where $c_1=\cos(\theta_1)$, $s_1=\sin(\theta_1)$, $s_{12}=\sin(\theta_1+\theta_2)$, and $c_{12}=\cos(\theta_1+\theta_2)$.
By taking derivative with respect to $t$ on both sides of the above equation, we have
\begin{equation}
	\begin{aligned}
		\dot{x}_1 &= -r_1 s_1 \dot{\theta}_1 -h_1 c_1 \dot{\theta}_1\\
		\dot{y}_1 &= r_1 c_1 \dot{\theta}_1 - h_1 s_1 \dot{\theta}_1\\
		\dot{x}_2 &= -l_1 s_1 \dot{\theta}_1 - (r_2 s_{12} + h_2 c_{12})(\dot{\theta}_1+\dot{\theta}_2)\\
		\dot{y}_2 &= l_1 c_1 \dot{\theta}_1 + (r_2 c_{12} - h_2 s_{12})(\dot{\theta}_1+\dot{\theta}_2).
	\end{aligned}
\end{equation}
Then, the kinetic energy $T$, potential energy $V$, and Lagrangian $L$ follow
\begin{equation}
	\begin{aligned}
		T&=\frac{1}{2}m_1(\dot{x}_1^2+\dot{y}_1^2)+\frac{1}{2}I_1\dot{\theta}_1^2+\frac{1}{2}m_2(\dot{x}_2^2+\dot{y}_2^2)+\frac{1}{2}I_2(\dot{\theta}_1+\dot{\theta}_2)^2\\
		V&=-m_1 g x_1 -m_2 g x_2 \\
		L&=T-V
	\end{aligned}
\end{equation}
where $g$ is the gravity constant. Note that the potential energy is parallel to the opposite of the gravity direction (if it has a direction). Based on Lagrange’s equation~\cite{bb20}, we have
\begin{equation}
	\begin{aligned}
		\frac{\partial}{\partial t}\frac{\partial L}{\partial \dot\theta_1}-\frac{\partial L}{\partial \theta_1}&=\tau_1,~
		\frac{\partial}{\partial t}\frac{\partial L}{\partial \dot\theta_2}-\frac{\partial L}{\partial \theta_2}&=\tau_2.
		\label{Lagr}
	\end{aligned}
\end{equation}
By taking derivative of $L$ with respect to $\theta_1$, $\theta_2$, $\dot\theta_1$, and $\dot\theta_2$, we obtain
\begin{equation}
	\begin{aligned}
		\frac{\partial L}{\partial \theta_1}&=-m_1 g (r_1 s_1 +h_1 c_1 ) - m_2 g (l_1 s_1 + r_2 s_{12}+h_2 c_{12})\\
		\frac{\partial L}{\partial \theta_2}&=  -m_2 \Big( (l_1 h_2 c_2+ l_1 r_2 s_2)\dot\theta_1^2 + (l_1 h_2 c_2 + l_1 r_2 s_2) \dot\theta_1 \dot \theta_2 + g r_2 s_{12} + g h_2 c_{12} \Big)\\
		\frac{\partial L}{\partial \dot\theta_1}
		&=(I_1 + I_2 + m_1 r_1^2 + m_2 r_2^2 + m_1 h_1^2 + m_2 h_2^2 + m_2 l_1^2 )\dot\theta_1 + (I_2+ m_2 r_2^2 +m_2 h_2^2 )\dot\theta_2 + (2 m_2 l_1 r_2 c_2 -2 m_2 l_1 h_2 s_2 )\dot\theta_1 \\
		&+(m_2 l_1 r_2 c_2 - m_2 l_1 h_2 s_2)\dot\theta_2\\
		\frac{\partial L}{\partial \dot\theta_2}&=(I_2+m_2 r_2^2 + m_2 h_2^2)\dot\theta_1 + (I_2 + m_2 r_2^2 + m_2 h_2^2)\dot\theta_2 + (m_2 l_1 r_2 c_2 - m_2 l_1 h_2 s_2)\dot\theta_1 
	\end{aligned}
	\label{derivative}
\end{equation} 
Moreover, we have
\begin{equation}
	\begin{aligned}
		&\frac{\partial}{\partial t}\frac{\partial L}{\partial \dot\theta_1}
		=(I_1 + I_2 + m_1 r_1^2 + m_2 r_2^2 + m_1 h_1^2 + m_2 h_2^2 + m_2 l_1^2 )\ddot\theta_1 + (I_2+m_2 r_2^2 +m_2 h_2^2 )\ddot\theta_2 + (2 m_2 l_1 r_2 c_2 -2 m_2 l_1 h_2 s_2 )\ddot\theta_1 \\
		&-(2 m_2 l_1 r_2 s_2 +2 m_2 l_1 h_2 c_2 )\dot\theta_1\dot\theta_2+(m_2 l_1 r_2 c_2 - m_2 l_1 h_2 s_2)\ddot\theta_2-(m_2 l_1 r_2 s_2 + m_2 l_1 h_2 c_2)\dot\theta_2^2\\
		&\frac{\partial}{\partial t}\frac{\partial L}{\partial \dot\theta_2}=(I_2+m_2 r_2^2 + m_2 h_2^2)\ddot\theta_1 + (I_2 + m_2 r_2^2 + m_2 h_2^2)\ddot\theta_2 + (m_2 l_1 r_2 c_2 - m_2 l_1 h_2 s_2)\ddot\theta_1 - (m_2 l_1 r_2 s_2 + m_2 l_1 h_2 c_2)\dot\theta_1\dot\theta_2
		\label{dtdl}
	\end{aligned}
\end{equation}

Substituting \eqref{derivative} and \eqref{dtdl} into \eqref{Lagr}, we obtain equation \eqref{dyn} where $\tau_1$ and $\tau_2$ are external torques:
\begin{equation}
	\begin{aligned}
		\tau_1 &= (I_1 + I_2 + m_1 r_1^2 + m_2 r_2^2 + m_1 h_1^2 + m_2 h_2^2 + m_2 l_1^2 )\ddot\theta_1 + (I_2+m_2 r_2^2 +m_2 h_2^2 )\ddot\theta_2 + (2 m_2 l_1 r_2 c_2 -2 m_2 l_1 h_2 s_2 )\ddot\theta_1 \\
		&-(2 m_2 l_1 r_2 s_2 +2 m_2 l_1 h_2 c_2 )\dot\theta_1\dot\theta_2+(m_2 l_1 r_2 c_2 - m_2 l_1 h_2 s_2)\ddot\theta_2-(m_2 l_1 r_2 s_2 + m_2 l_1 h_2 c_2)\dot\theta_2^2+m_1 g (r_1 s_1 +h_1 c_1 ) \\
		& +m_2 g (l_1 s_1 + r_2 s_{12}+h_2 c_{12})\\
		\tau_2&=(I_2+m_2 r_2^2 + m_2 h_2^2)\ddot\theta_1 + (I_2 + m_2 r_2^2 + m_2 h_2^2)\ddot\theta_2 + (m_2 l_1 r_2 c_2 - m_2 l_1 h_2 s_2)\ddot\theta_1+ m_2(l_1 h_2 c_2+ l_1 r_2 s_2)\dot\theta_1^2 \\& + m_2 g r_2 s_{12} + m_2 g h_2 c_{12} 
		\label{dyn}
	\end{aligned}
\end{equation}
Using the following parameter mapping
\begin{equation}
	\begin{aligned}
		&X_2= m_2 r_2, Y_2= m_2 h_2,X_1 =m_1 r_1 + m_2 l_1, Y_1 = m_1 h_1 \\
		& J_2 = I_2 + m_2 (r_2^2+ h_2^2), J_1 = J_2 + I_1 + m_1(r_1^2+h_1^2)+ m_2 l_1^2
		\label{spy}
	\end{aligned}
\end{equation}
and substituting \eqref{spy} into \eqref{dyn}, we obtain
\begin{equation}
	\begin{aligned}
		\tau_1 &= \big(J_1+ 2 l_1 (X_2 c_2 -Y_2 s_2)\big) \ddot\theta_1 + (J_2+ l_1 (X_2 c_2 -Y_2 s_2))\ddot\theta_2 -l_1 (X_2 s_2 + Y_2 c_2)\big(2\dot\theta_1 \dot\theta_2 +\dot\theta_2^2 \big)\\
		&+g(X_1 s_1+ Y_1 c_1 + X_2 s_{12}+ Y_2 c_{12})\\
		\tau_2&=\big(J_2 + l_1 (X_2 c_2 - Y_2 s_2)\big)\ddot\theta_1 + J_2 \ddot\theta_2+l_1(X_2 s_2+ Y_2 c_2)\dot\theta_1^2 +g(X_2 s_{12}+ Y_2 c_{12})
	\end{aligned}
\end{equation}
Furthermore, we denote
\begin{equation}
	\begin{aligned}
		M_{11}&=J_1+ 2 l_1 (X_2 c_2 -Y_2 s_2), M_{12}=J_2+ l_1 (X_2 c_2 -Y_2 s_2)\\
		M_{21}&=J_2 + l_1 (X_2 c_2 - Y_2 s_2), ~M_{22}=J_2\\
		C_{11}&=-2l_1 (X_2 s_2 + Y_2 c_2)\dot\theta_2, ~C_{12}=-l_1 (X_2 s_2 + Y_2 c_2)\dot\theta_2\\
		C_{21}&=l_1(X_2 s_2+ Y_2 c_2)\dot\theta_1, ~~~~C_{22}=0\\
		G_{1}&=g(X_1 s_1+ Y_1 c_1 + X_2 s_{12}+ Y_2 c_{12})\\
		G_{2}&=g(X_2 s_{12}+ Y_2 c_{12}).
		\label{MCG}
	\end{aligned}
\end{equation}
Finally, the dynamics of the two-link robots can be written as
\begin{equation}
	\tau=M(\theta)\begin{bmatrix}
		\ddot\theta_1\\
		\ddot\theta_2
	\end{bmatrix}+C(\theta,\dot{\theta})\begin{bmatrix}
		\dot\theta_1\\
		\dot\theta_2
	\end{bmatrix}+G(\theta)
\end{equation}
where $M(\theta)=\begin{bmatrix}
	M_{11},M_{12}\\
	M_{21},M_{22}
\end{bmatrix}$, $C(\theta,\dot{\theta})=\begin{bmatrix}
	C_{11},C_{12}\\
	C_{21},C_{22}
\end{bmatrix}$, $G(\theta)=\begin{bmatrix}
	G_1\\
	G_{2}
\end{bmatrix}$, and $\tau=\begin{bmatrix}
	\tau_1\\
	\tau_2
\end{bmatrix}$. It is worth mentioning that the external torques are composed of the motor torque $\tau_{mot}$ and the friction $\tau_{fri}$. Due to the characteristic of harmonic reducer, we employ the Coulomb-viscous-Stribeck friction model:
\begin{equation}
	\tau_{fri}= \big(\tau_c+(\tau_s-\tau_c)\exp(-\dot\theta/\dot\theta_{s})\big)sgn(\dot\theta)+ \eta \dot\theta
\end{equation}
where $\tau_c$ and $\tau_s$ are Coulomb friction and static friction torque, $\dot\theta_{s}$ is the Stribeck angular rate, and $\eta$ is the viscous friction coefficient. Thus, we have
\begin{equation}
	\begin{aligned}
		\tau_1&=\tau_{mot,1}-\tau_{fri,1}\\
		\tau_2&=\tau_{mot,2}-\tau_{fri,2}.
	\end{aligned}
\end{equation}

\begin{table*}[htbp]
	\centering
	\caption{Exoskeleton Parameters.}
	\scalebox{0.8}{
		\begin{tabular}{llllllll}
			\hline
			\hline
			\multirow{2}{*}{Knee}&$\tau_{c,2}$& $\tau_{s,2}$ & $\dot{q}_{s,2}$ & $\eta_{2}$ &  $X_2$& $Y_2$ & $J_2$ \\
			&2.582 & 6.216 & 2.886 & 6.495 & 0.592 & 0.01 & 0.549\\
			\multirow{2}{*}{Hip}&$\tau_{c,1}$& $\tau_{s,1}$ & $\dot{q}_{s,1}$ & $\eta_{1}$ &  $X_1$& $Y_1$ & $J_1$ \\
			&9.964 & 6.141 & 19.311 & 3.967  & 3.746 & 0.01 & 1.671\\
			\hline
			\hline
	\end{tabular}}
	\label{leg}
\end{table*}

We identify the system parameters using the human gait as commanded position at the frequencies from $0.1-0.5$ Hz with the increment of 0.1 Hz where the optimization algorithm is the simulated annealing optimization (MATLAB 2019b optimization toolbox). The results are summarized in Table. \ref{leg}.

\subsection{Nonlinear Disturbance Observer}
According to Section \ref{section1} and considering the unknown disturbance, the exoskeleton model has
\begin{equation}
	M(\theta)\ddot{\theta} + C(\theta,\dot{\theta})\dot{\theta} + G(\theta) = \tau + d 
	\label{twolink}
\end{equation}
where $M(\theta) \in \mathbb{R}^{2}$ is the positive definite inertial matrix, $C(\theta,\dot{\theta}) \in \mathbb{R}^{2}$ is the Coriolis matrix, and $G(\theta) \in \mathbb{R}^{2}$ is the gravity matrix. Subsequently, we have
\begin{equation}
	d=M(\theta)\ddot{\theta}+C(\theta,\dot{\theta})\dot{\theta}+G(q)-\tau.
	\label{raw}
\end{equation}
By assuming that $\dot{d}=0$, one can estimate the disturbance as
\begin{equation}
	\dot{\hat{d}}= - L (\theta,\dot{\theta}) (\hat{d}-d)
\end{equation}
where $L (\theta,\dot{\theta})$ is a term to be determined. Define $e=d-\hat{d}$ and $\dot{e}=\dot{d}-\dot{\hat{d}}$, we obtain
\begin{equation}
	\dot{e}+L (\theta,\dot{\theta})e=0.
\end{equation}
In this case, the stability of the observer can be guaranteed by setting $L (\theta,\dot{\theta})$ carefully. 

To proceed, define auxiliary vector $z=\hat{d}- p(\theta,\dot{\theta})$ where $p(\theta,\dot{\theta})$ is to be determined. Let $L(\theta,\dot{\theta})$ follows 
\begin{equation}
	L(\theta,\dot{\theta})M(\theta)\ddot{\theta} = \begin{bmatrix}\frac{\partial p(\theta,\dot{\theta})}{\partial \theta}, \frac{\partial p(\theta,\dot{\theta})}{\partial \dot{\theta}}\end{bmatrix}\begin{bmatrix}
		\dot{\theta}\\
		\ddot{\theta}
	\end{bmatrix}.
\end{equation}
Then, we have
\begin{equation}
	\begin{aligned}
		&\dot{z}= \dot{\hat{d}}-\frac{p(\theta,\dot{\theta})}{t}\\
		&=\dot{\hat{d}} -\begin{bmatrix}\frac{\partial p(\theta,\dot{\theta})}{\partial \theta}, \frac{\partial p(\theta,\dot{\theta})}{\partial \dot{\theta}}\end{bmatrix}\begin{bmatrix}
			\dot{\theta}\\
			\ddot{\theta}
		\end{bmatrix}\\
		&=- L (\theta,\dot{\theta}) (\hat{d}-d) - L(\theta,\dot{\theta})M(\theta)\ddot{\theta}\\
		&=- L (\theta,\dot{\theta}) \Big(z+ p(\theta,\dot{\theta})-C(\theta,\dot{\theta})\dot{\theta}-G(\theta)+\tau\Big)\\
		&= -L (\theta,\dot{\theta})z+ L (\theta,\dot{\theta})\Big( C(\theta,\dot{\theta})\dot{\theta}+G(\theta)-\tau-p(\theta,\dot{\theta})\Big).
	\end{aligned}
\end{equation} 
It follows that $\hat{d}=z+p(\theta,\dot{\theta})$.
The stability of this observer can be guaranteed by selecting $L(\theta,\dot{\theta})$ carefully since 
\begin{equation}
	\begin{aligned}
		\dot{e}&= \dot{d}- \dot{\hat{d}}=-\dot{z}- \frac{\partial p(\theta,\dot{\theta})}{\partial t}\\
		&=L (\theta,\dot{\theta}) \Big(z+ p(\theta,\dot{\theta})-C(\theta,\dot{\theta})\dot{\theta}-G(\theta)+\tau -M(\theta)\ddot{\theta}\Big)\\
		&=L (\theta,\dot{\theta})\Big(z+p(\theta,\dot{\theta})-d\Big)\\
		&=-L (\theta,\dot{\theta})e.
	\end{aligned}
\end{equation}
For two-link robots, we can select~\cite{bb5}
\begin{equation}
	p(\theta,\dot{\theta})=c\begin{bmatrix}
		\dot{\theta}_1\\
		\dot{\theta}_1 +\dot{\theta}_2
	\end{bmatrix}
\end{equation}
where $c$ should be bigger than a certain level. By this setting, $\frac{\partial p(\theta,\dot{\theta})}{\partial t}= \begin{bmatrix}\frac{\partial p(\theta,\dot{\theta})}{\partial \theta}, \frac{\partial p(\theta,\dot{\theta})}{\partial \dot{\theta}}\end{bmatrix}\begin{bmatrix}
	\dot{q}\\
	\ddot{\theta}
\end{bmatrix}=c\begin{bmatrix}
	1,0 \\
	1,1 
\end{bmatrix}\ddot{\theta}$ and $L(\theta,\dot{\theta})=c\begin{bmatrix}
	1,0 \\
	1,1 
\end{bmatrix}M(\theta)^{-1}$. The continuous nonlinear observer can be written as 
\begin{equation}
	\begin{aligned}
		\dot{z}&=-L (\theta,\dot{\theta})z+ L (\theta,\dot{\theta})\Big( C(\theta,\dot{\theta})\dot{q}+G(\theta)-T-p(\theta,\dot{\theta})\Big)\\
		\hat{d}&=z+p(\theta,\dot{\theta}).
		\label{NDOB}
	\end{aligned}
\end{equation}
Based on Euler discretization, the discrete form of the observer can be written as 
\begin{equation}
	\begin{aligned}
		z_{k+1} &= z_k+ \Big(-Lz_k+ L\big( C\dot{\theta}+G-\tau-p\big)\Big)\Delta T\\
		&= (I- L \Delta T) z_k + L\Delta T\big( C\dot{\theta}+G-\tau-p\big)\\
		\hat{D}&=z_k + p
	\end{aligned}
\end{equation}
where $L \triangleq L(\theta_k,\dot{\theta}_k)$, $C \triangleq C(\theta_k,\dot{\theta}_k)$, $G \triangleq G(\theta_k)$, and $p \triangleq p(\theta_k,\dot{\theta}_k)$.
\color{black}
\subsection{Influence of Observer Parameters}
\begin{figure}
	\centering
	\includegraphics[width=0.25\linewidth]{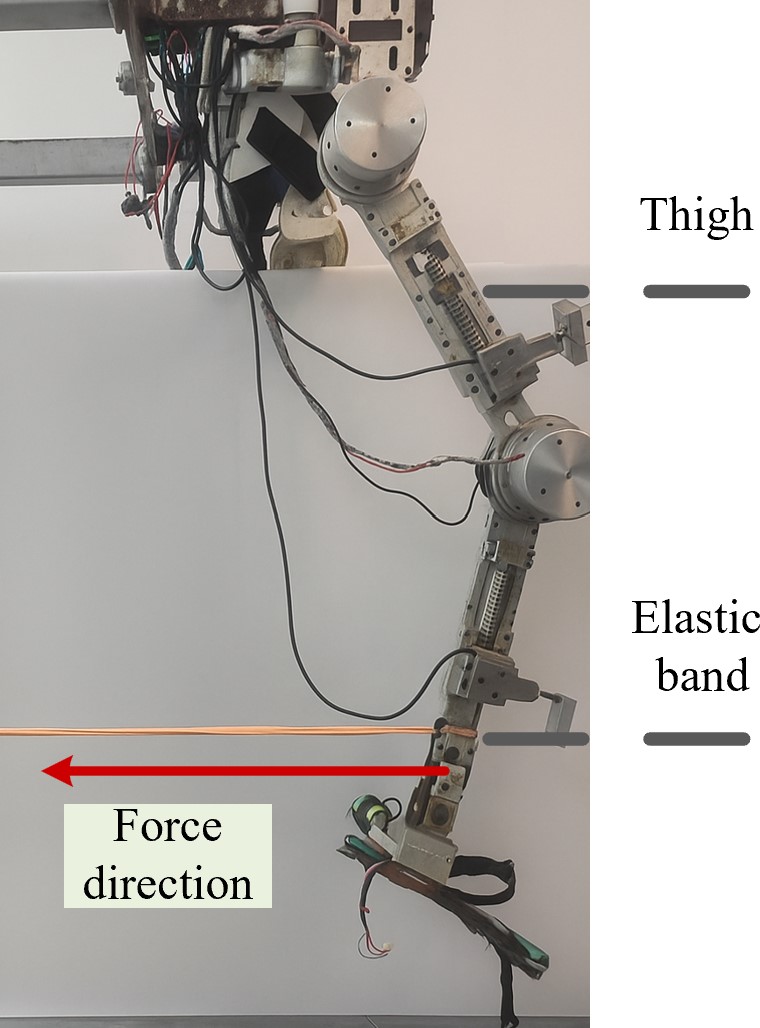}
	\caption{ Sweeping parameter experiment with gait frequency $f=0.3$ Hz. To simulate the dynamic disturbance, an elastic band is attached to the shank of the exoskeleton, with its other end fixed to a stationary frame. In experiments, the band is stretched to varying lengths in sync with the gait trajectory to simulate different levels of disturbance. The maximum allowable force generated by the elastic band is 5 kg.}    \label{sweep}
\end{figure}

To investigate the influence of the Markov transition probability matrix in IMMEKF-DOB and the effects of kernel bandwidth $\sigma_d$ in dynamic disturbance environment, we conducted two experiments by comparing the tracking errors of different observers by varying the Markov transition probability matrix or the disturbance kernel bandwidths. The experimental setup is shown in Fig. \ref{sweep}. 

\begin{figure}[htbp]
	\centering
	\subfigure[]{\includegraphics[width=3in]{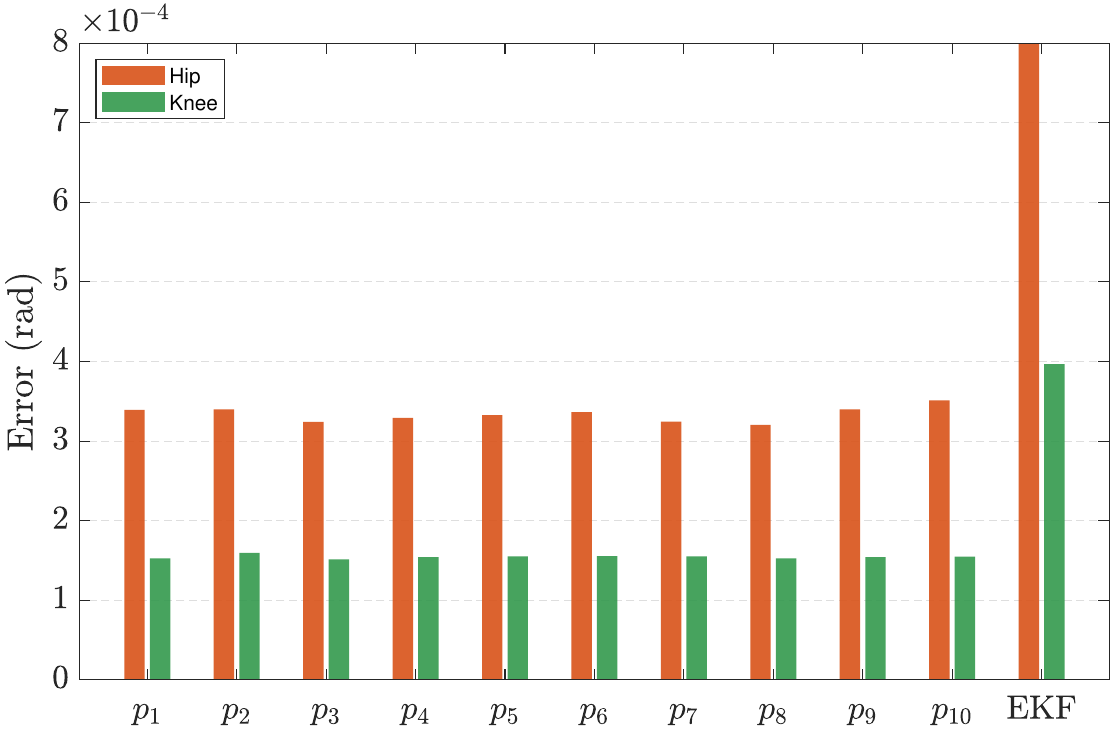}
		\label{psweep}}
	\subfigure[]{\includegraphics[width=3in]{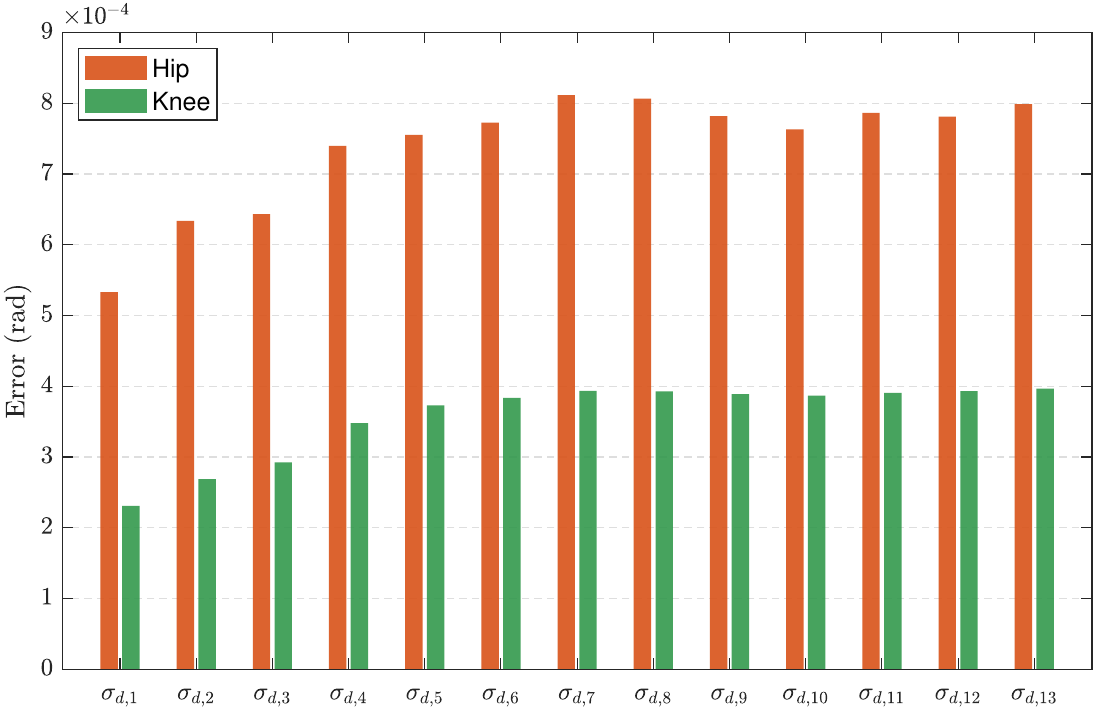}
		\label{sigmasweep}}
	\caption{Tracking errors in sweeping experiments. (a) The tracking error of IMMEKF-DOB with varying $p$ and its comparison with EKF. (b) The tracking error of MKCEKF-DOB with varying $\sigma_d$.}		
	\label{sweepexp}
\end{figure}

In the probability matrix sweeping experiment, we set the adjustable  parameter as follows:
\begin{equation}
	P=\begin{bmatrix}
		p&1-p\\
		1-p&p
	\end{bmatrix}
\end{equation}
where $p$ varies from 0.05 to 0.95 at an increment of 0.05 and is denoted as $p_{j}$ where $j\in [1,10]$, respectively. Be aligned with our previous experiments, two different disturbance covariances, i.e., $Q_{d1}=Q_{d}$ and $Q_{d2}=20Q_{d}$, are used. The corresponding tracking error of applying different $p_j$ is visualized in Fig. \ref{psweep}. It can be observed that the tracking errors are not very sensitive to the selection of $p$, which is beneficial to the practical implementation of IMMEKF-DOB. We also observe that our developed IMMEKF-DOB outperforms the conventional EKF-DOB significantly in all situations, indicating the proposed observer is suitable for situations with complex disturbances.

We further conducted experiments to investigate the effects of $\sigma_d = \exp(\eta_j)$ where $\eta_j$ varies from $-1$ to $5$ at an increment of 0.5, i.e., $\sigma_d \in [\exp(-1), \exp(5)]$. The corresponding results are visualized in Fig. \ref{sigmasweep} where $\sigma_{d,1}$ denotes the results of applying $\sigma_d= \exp(-1)$ and $\sigma_{d,13}$ denotes the results of applying $\sigma_d= \exp(5)$. According to Remark 3 of the main manuscript, the proposed MKCEKF-DOB is identical to EKF-DOB by applying a large enough kernel bandwidth, which indicates that the tracking results of $\sigma_{d,13}$ can be regarded as the result of EKF-DOB. Under this substitution, we observe that the MCKEFK-DOB is always better or equivalent to the EKF-DOB (if we ignore the tiny error caused by some random factors in different experiments). Moreover, we find that a smaller kernel bandwidth is much more appreciated for scenarios with dynamically changed disturbances, since it can inflate the disturbance error covariance effectively.  
\subsection{Simulations on Manipulators}
\begin{figure}[h]
	\centering 
	\subfigure[Initial state]{
		\includegraphics[width=0.43\textwidth]{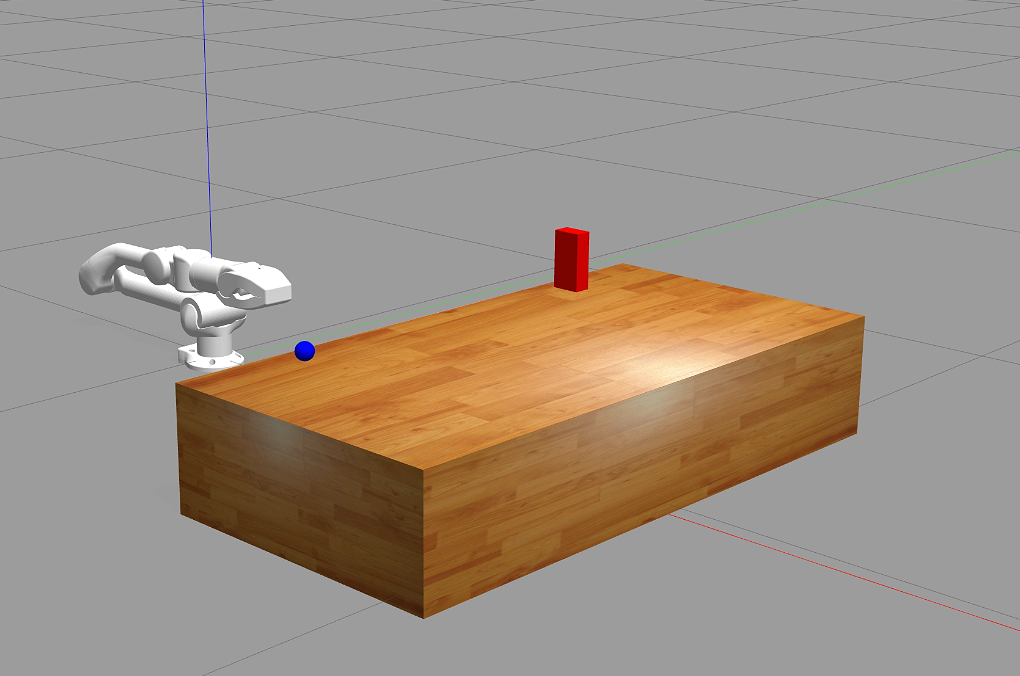}  
	}
	\subfigure[grasping state]{
		\includegraphics[width=0.45\textwidth]{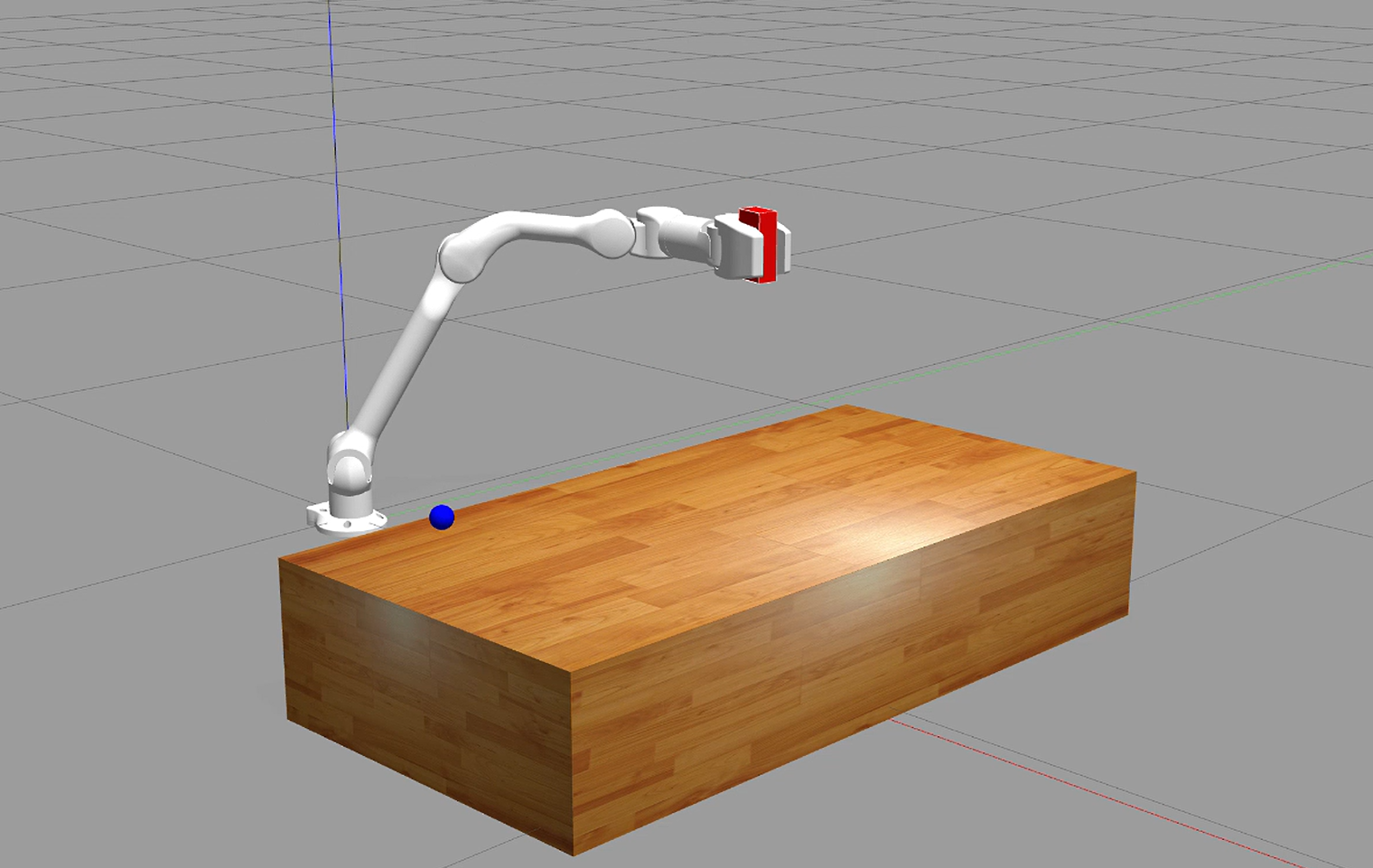} 
	}
	\caption{Simulation scenarios.} 
	\label{fig:gazebo}
\end{figure}
\color{black}
	We test our algorithms on the Unitree Z1 pro six-axis manipulator in the Gazebo simulation environment. The experimental scenario is shown in Fig. \ref{fig:gazebo}, where a pick-and-place task is used and the disturbance comes from the frictions and unknown payload. In simulations, the ground truth weight of the cube is 1 kg with size 5 cm $\times$ 3 cm $\times$ 10 cm. The initial cube position is ${p}_{cube}=(0.6, 0.3, 0.25)$ and the desired cube position is ${p}_{des}=(0.6, 0.1, 0.25)$. In simulations, the robotic arm starts from its initial position, grasps the object and places it at the desired position, and finally goes back to its initial position. The trajectory is generated through the minimum jerk approach to reduce the mechanical shock of the manipulator. The controller operates at 200Hz and uses the following PD control law:
	\begin{equation}
		\tau=25.6K_p(q_d-q)+0.0128K_d(\dot{q_d}-\dot{q})+\tau_{f},
	\end{equation}
	where $K_p =\operatorname{diag}{[20, 30, 30, 20, 15, 10]}$ and $K_d =\operatorname{diag}{[2000, 2000, 2000, 2000, 200, 300]}$, $\tau_{f}$ is the feedforward torque obtained by the manipulator dynamics using \textit{Pinocchio} Python library.
	
\begin{figure}[h] 
		\centering 
		\includegraphics[width=0.8\textwidth]{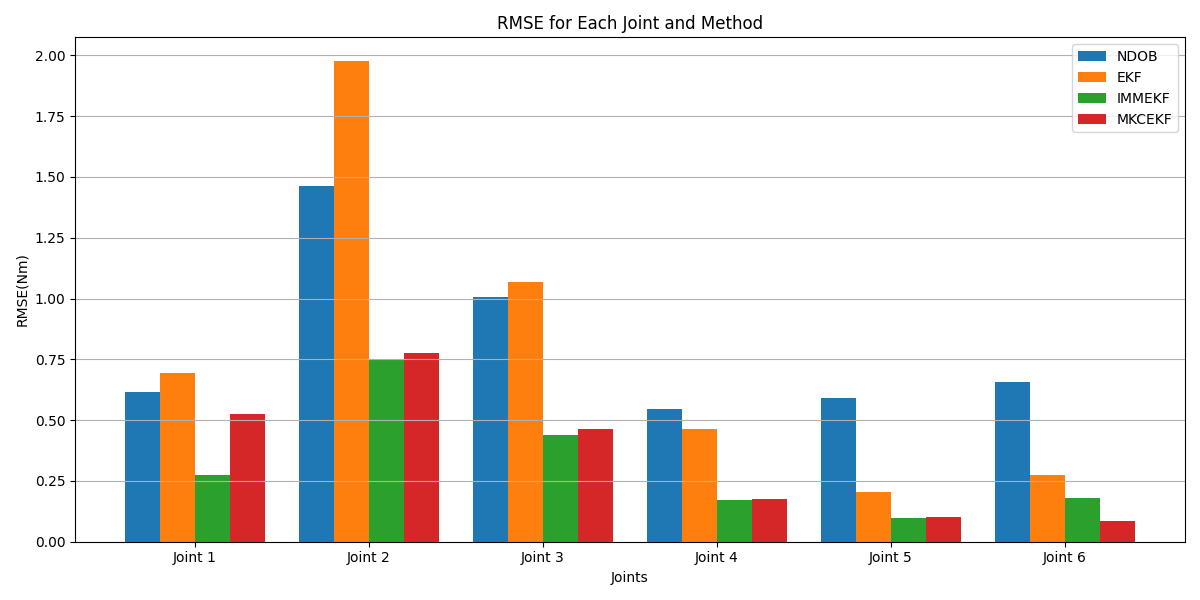} 
		\caption{RMSE comparison under 1 kg load.} 
		\label{fig:1kg RMSE} 
	\end{figure}

\begin{figure}[h] 
		\centering 
		\includegraphics[width=0.8\textwidth]{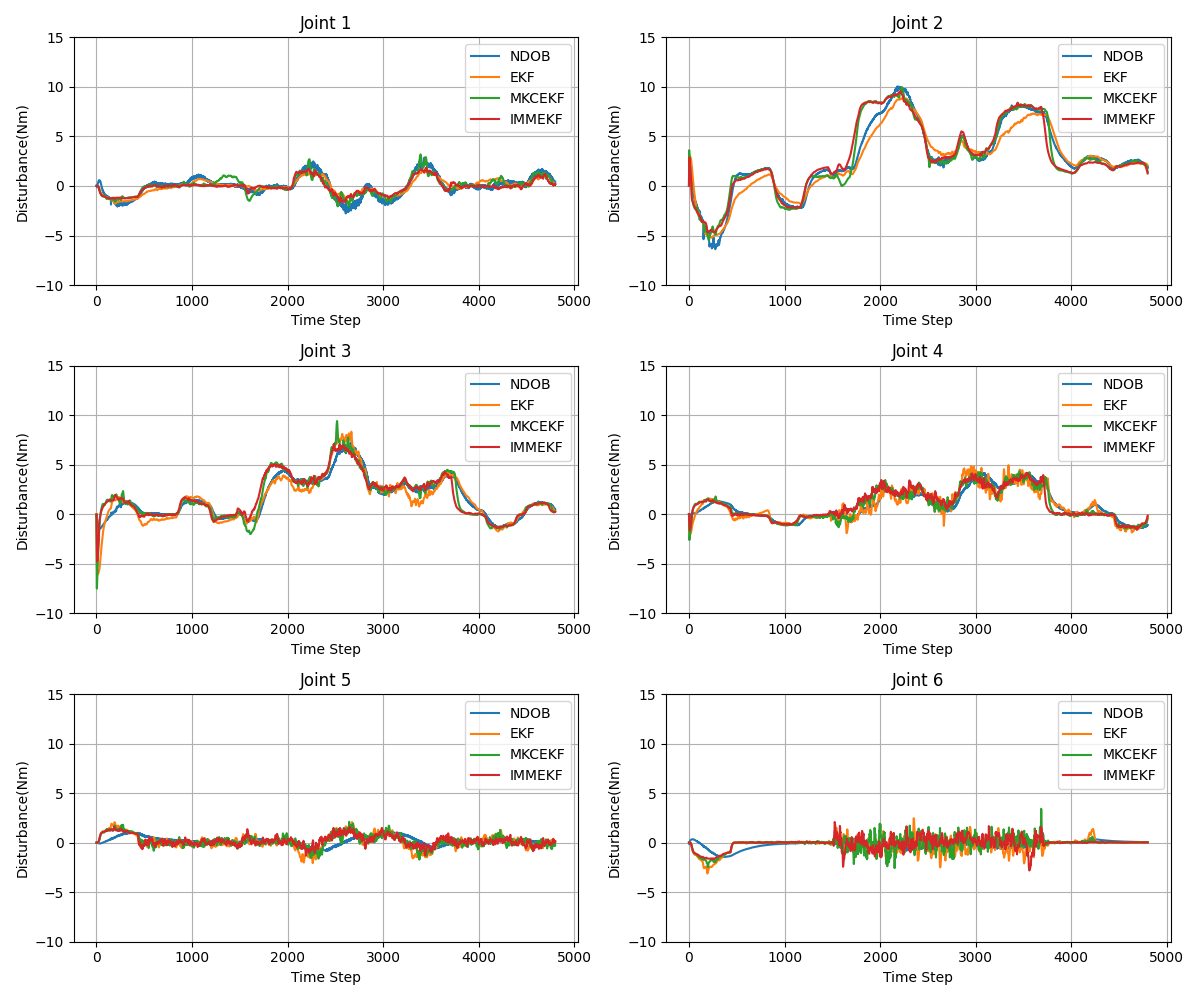} 
		\caption{Disturbance estimation performance under 1kg load.} 
		\label{fig:1kg load} 
	\end{figure}

	In EKF, the state vector has \(x = [d, \dot{q}, q]\) and the process covariance and measurement covariance have
	\begin{equation}
		Q_{ekf} = \begin{bmatrix}
			10^{-3}I_6 & 0 & 0 \\
			0 & 10^{-3}I_6 & 0 \\
			0 & 0 & 10^{-4}I_6
		\end{bmatrix},
		R_{ekf} = \begin{bmatrix}
			10^{-2}I_6 & 0 \\
			0 & 10^{-3}I_6
		\end{bmatrix}.
	\end{equation}
	In MKCEKF, we use the same covariance matrices, along with the kernel bandwidth vector as follows:
	\begin{equation}
		\sigma_p =[0.5\mathbf{1}_6, 10^2 \mathbf{1}_{12}],\sigma_r = 10^4 \mathbf{1}_{12},
	\end{equation}
	where $\mathbf{1}_n$ is a vector which all n-dimensional elements are 1. In IMMEKF, two sets of covariance matrices are selected, where
	\begin{equation}
		Q_{k1}=Q_{ekf},Q_{k2}=\begin{bmatrix}
			5*10^{-1}I_6 & 0 & 0 \\
			0 & 10^{-3}I_6 & 0 \\
			0 & 0 & 10^{-4}I_6
		\end{bmatrix}
	\end{equation}
	\begin{equation}
		R_{k1}=R_{k2}=\begin{bmatrix}
			10^{-2}I_6 & 0 \\
			0 & 10^{-3}I_6
		\end{bmatrix}.
	\end{equation}
	We choose the probability distribution matrix as
	\begin{equation}
		T=\begin{bmatrix}
			0.95 & 0.05\\
			0.15 & 0.85\\
		\end{bmatrix}
	\end{equation}
	We tested the algorithms in the above set using the same experimental setup. The root mean squared error (RMSE) of the estimated disturbances for the six joints is shown in Fig. \ref{fig:1kg RMSE}, and the corresponding disturbance estimation curves are presented in Fig. \ref{fig:1kg load}. We observe that the disturbance estimation accuracy of IMMEKF and MKCEKF improves by $62.18\%$ and $60.68\%$ for joint 2, $59.08\%$ and $56.65\%$ for joint 3, and $62.97\%$ and $61.71\%$ for joint 4, respectively.
\clearpage
\bibliographystyle{IEEEtran}
\bibliography{reference}
	
\end{document}